\documentclass[10pt,twocolumn,letterpaper]{article}

\usepackage{cvpr}              
\usepackage[accsupp]{axessibility}
\usepackage{wrapfig}
\usepackage{amssymb}
\usepackage{makecell}
\usepackage{multirow}

\definecolor{cvprblue}{rgb}{0.21,0.49,0.74}
\usepackage[pagebackref,breaklinks,colorlinks,allcolors=cvprblue]{hyperref}

\def\paperID{14592} 
\def\confName{CVPR}
\def\confYear{2025}

\title{Six-CD: Benchmarking Concept Removals for Text-to-image Diffusion Models}

\author{Jie Ren$^{1}$\thanks{Equal contribution.}, Kangrui Chen$^{1 *}$, Yingqian Cui$^{1}$, Shenglai Zeng$^{1}$, Hui Liu$^{1}$, Yue Xing$^{1}$, \\
Jiliang Tang$^{1}$, Lingjuan Lyu$^{2}$\\
$^{1}$Michigan State University ~~~~~   $^{2}$Sony AI\\
{\tt\small \{renjie3,chenkan4,cuiyingq,zengshe1,liuhui7,xingyue1,tangjili\}@msu.edu} \\
{\tt\small lingjuan.lv@sony.com}
}

\begin{document}
\maketitle
\begin{abstract}
Text-to-image (T2I) diffusion models have shown exceptional capabilities in generating images that closely correspond to textual prompts. However, the advancement of T2I diffusion models presents significant risks, as the models could be exploited for malicious purposes, such as generating images with violence or nudity, or creating unauthorized portraits of public figures in inappropriate contexts. To mitigate these risks, concept removal methods have been proposed. These methods aim to modify diffusion models to prevent the generation of malicious and unwanted concepts. Despite these efforts, existing research faces several challenges: (1) a lack of consistent comparisons on a comprehensive dataset, (2) ineffective prompts in harmful and nudity concepts, (3) overlooked evaluation of the ability to generate the benign part within prompts containing malicious concepts. To address these gaps, we propose to benchmark the concept removal methods by introducing a new dataset, Six-CD, along with a novel evaluation metric. In this benchmark, we conduct a thorough evaluation of concept removals, with the experimental observations and discussions offering valuable insights in the field.
\end{abstract}  
\section{Introduction}
\label{sec:intro}

Diffusion models have demonstrated remarkable capabilities in image generation~\cite{ho2020denoising,song2020score,ho2022classifier,song2020denoising,dhariwal2021diffusion}, particularly text-to-image (T2I) diffusion models~\cite{stablediffusion2, chen2023pixart, ramesh2022hierarchical}. These models have gained widespread popularity and deployment due to their ability to generate images that precisely align with textual prompts. However, T2I diffusion models can be exploited for malicious purposes, such as creating images depicting violence, nudity, or fake celebrity images in undesirable contexts like jail~\cite{rando2022red,washingtonpost2023,dalle2systemcard2023, somepalli2023diffusion,shan2023glaze,ren2024copyright, li2024neural}. To ensure the integrity of the model developers, it is crucial to design \textit{benign} models that censor malicious concepts and produce only safe and appropriate content.

Strategies have been proposed to mitigate the malicious generation. For example, in the data collection stage, Stable Diffusion (SD)~\cite{stablediffusion2} and Adobe~\cite{rao2023adobe} filter out the {Not-Safe-For-Work (NSFW)} contents in training data. After image generation, SD uses a detector to ensure the images are benign, removing unwanted ones. However, training data filtering requires retraining and will be inefficient if the model has already been trained on various data. For open-sourced models like SD~\cite{stablediffusion2}, 
the detector {can be easily disabled by modifying the source code}. Therefore, concept removal techniques have been proposed to edit a trained model's parameters to prevent generating malicious concepts~\cite{gandikota2023erasing, kumari2023ablating, zhang2023forget, ren2025general, li2021online}. For instance, methods are proposed to fine-tune the models to replace unwanted concepts with benign alternatives~\cite{gandikota2023erasing, kumari2023ablating, zhang2023forget}. 
Besides, to accelerate fine-tuning, some methods fine-tune the linear components (e.g. the weight matrix of cross attention) using the closed-form optimal solution~\cite{gandikota2024unified, lu2024mace, xiong2024editing}. {Alternatively, } there are methods modifying the output during inference to bypass fine-tuning~\cite{schramowski2023safe, brack2024sega}.

\begin{table*}[t]
    \centering
   \caption{Categories of unwanted concepts in existing literature}
    \resizebox{1\textwidth}{!}{ 
    \begin{tabular}{lccccccccccc}
        \toprule
        & ESD~\cite{gandikota2023erasing} & SPM~\cite{lyu2023one} & SDD~\cite{kim2023towards} & FMN~\cite{zhang2023forget} & UCE~\cite{gandikota2024unified} & MACE~\cite{lu2024mace} & EMCID~\cite{xiong2024editing} & SLD~\cite{schramowski2023safe} & SEGA~\cite{brack2024sega} & UC~\cite{zhang2024unlearncanvas} & CPDM~\cite{ma2024dataset} \\
        \midrule
        Harmful  &  &  & \checkmark &  &  &  &  & \checkmark & \checkmark &  &   \\
        Nudity & \checkmark & \checkmark & \checkmark &  & \checkmark & \checkmark & \checkmark & \checkmark & \checkmark &  &   \\
        Celebrity &  &  &  & \checkmark &  & \checkmark &  &  &  &  & \checkmark \\
        Copyrighted &  & \checkmark &  &  &  &  &  &  &  &  & \checkmark\\
        Object & \checkmark &  & \checkmark & \checkmark & \checkmark & \checkmark & \checkmark &  & \checkmark & \checkmark &   \\
        art style & \checkmark & \checkmark & \checkmark & \checkmark & \checkmark & \checkmark & \checkmark &  & \checkmark & \checkmark & \checkmark\\
        \bottomrule
    \end{tabular}
    }
    \label{tab:category}
\end{table*}
While there is rich literature improving concept removal methods, we observe three potential issues:
\begin{itemize}[leftmargin=0.6cm]
    \item[1.] \textbf{Lack of consistent and comprehensive comparisons.} Current concept removal methods, as shown in Table~\ref{tab:category}, typically analyze limited categories, lacking consistent and comprehensive comparisons. This gap prevents a thorough understanding of the methods' behavior.
    \item[2.] \textbf{Ineffective prompts.} In existing datasets~\cite{schramowski2023safe}, some categories, such as nudity, contain a large portion of ``ineffective prompts'', which only trigger the model to generate malicious contents with a low probability (see Fig.~\ref{fig:i2p_bar}). This can result in inefficient evaluation since the ineffective prompts will generate many benign images, and evaluating concept removal on them is meaningless. Meanwhile, in other categories of unwanted concepts like celebrities, the prompts are more effective than the nudity concept (see Fig.~\ref{fig:i2p_bar} and Fig.~\ref{fig:main}). When the builder has to remove both nudity and celebrity concepts, the ineffective prompts may cause unfair comparisons. The effective prompts in celebrity can generate more malicious images, which may falsely induce the builder to put more weight on celebrity concepts.
    \item[3.] \textbf{Lack of evaluation on in-prompt retainability.} Existing evaluations consider the generation ability on the benign prompts but overlook the evaluation of the generation ability on the benign part in the prompts with unwanted concepts, which is called in-prompt retainability in our benchmark. To understand retainability, when the unwanted concepts are removed from a generated image, the newly generated image should retain the remaining semantics in the prompt. For example, if we remove ``\textit{Mickey Mouse}'' from the prompt of ``\textit{Mickey Mouse is eating a burger}'', the generated image should still depict ``\textit{eating a burger}''. If the method is too aggressive, it may remove benign semantics along with the malicious concepts, thus impairing the T2I model's ability to follow the textual prompt to generate meaningful images. Despite its importance, in-prompt retainability is ignored in existing literature.
\end{itemize}

To tackle the aforementioned problems, we aim to benchmark concept removal methods by introducing a comprehensive dataset and a new evaluation metric. Our contributions are as follows: 

\begin{itemize}[leftmargin=0.6cm]
    \item[1.] \textbf{A comprehensive dataset.} We propose a new dataset, Six-CD, which contains \textbf{Six} categories of unwanted \textbf{C}oncepts in \textbf{D}iffusion models, including \textit{harm, nudity, identities of celebrities, copyrighted characters, objects,} and \textit{art styles}. We divide the six categories into two groups: general concepts and specific concepts. General concepts, such as harm and nudity, are of concern to all users. Specific concepts {are subject} to a specific {entity, e.g.,} a person or a company. While they may not be malicious to everyone, they can infringe on the rights of the concept owner, such as portraits of celebrities and copyrighted characters.
    \item[2.] \textbf{Effective prompts.} Through extensive experiments in Sec.~\ref{sec:ineffective} and Sec.~\ref{exp:main}, we observe that ineffective prompts happen more frequently in general concepts. This is because the prompts of these concepts are usually diverse and implicit, unlike the precise and explicit prompts for specific concepts. Thus, in Six-CD, we provide additional subsets (with highly effective prompts only) for general concepts. Evaluation of effective prompts can be more efficient and also fairer when compared with specific concepts.
    \item[3.] \textbf{A new evaluation metric.} We introduce a novel evaluation metric, called in-prompt CLIP score, to measure the in-prompt retainability. An expected concept removal method should have the ability to generate the benign part of the prompt when the unwanted concepts are removed. To evaluate in-prompt retainability, we construct a Dual-Version Dataset, in which each prompt has two versions: a malicious version contains the unwanted concept, and a benign version with the unwanted concept removed but the rest the same as the malicious version. 
    We apply concept removal methods to generate images from the malicious version and then use the CLIP score~\cite{hessel2021clipscore} to measure the similarity between these images and the benign prompts. Ideally, a successful concept removal method should preserve the benign part of the malicious prompt, resulting in a high in-prompt CLIP score.\vspace{-0.05in}
\end{itemize}

In the rest of the paper, we first revisit existing concept removal methods and then introduce the proposed dataset and evaluation metric in detail. Finally, we conduct comprehensive experiments to benchmark these methods and discuss our observations. 
\section{Preliminaries}
\subsection{Text-to-image Diffusion Models} 
Diffusion models typically involve a \textit{forward} diffusion process and its \textit{reverse} diffusion process. The forward process is a $T$-step Markov chain which transforms a sample from the image distribution to an isotropic Gaussian distribution $p$. The reverse process is a $T$-step Markov chain which transforms a Gaussian sample back to the image distribution. 
The reverse process can be represented as 
\begin{align} \label{eq:markov}
    p_\theta\left(x_{0: T}\right) & =p\left(x_T\right) \textstyle\prod_{t=1}^T p_\theta\left(x_{t-1} \mid x_t\right) \\
    p_\theta\left(x_{t-1} \mid x_t\right) & =\mathcal{N}\left(x_{t-1} ; \mu_\theta\left(x_t, t\right), \Sigma_\theta\left(x_t, t\right)\right),
    \label{eq:reverse}
\end{align}
where $x_0$ is the image sample, $x_T$ is the Gaussian sample, and $x_t (0<t<T)$ is the middle state of diffusion. The terms $\mu_\theta$ and $\Sigma_\theta$ are estimated by a denoising network $\epsilon_\theta (x_t, t)$ which takes $x_t$ and $t$ as input. 
By $\epsilon_\theta$, an image sample $x_0$ can be generated from a Gaussian sample $x_T$ following Eq.~\eqref{eq:markov}.

Text-to-image (T2I) diffusion models further extend the diffusion process by guiding it with a textual prompt $y$~\cite{rombach2022high, deepif, chen2023pixart}. Specifically, $\epsilon_\theta (x_t, t)$ is modified to take $y$ as an additional input, resulting in $\epsilon_\theta (x_t, y, t)$. The network $\epsilon_\theta (x_t, y, t)$ uses a cross-attention module to select information from the textual prompt to guide the diffusion process. In cross attention, each token is transformed into key ($K$) space and value ($V$) space, while the image is transformed into query ($Q$) space. The key and the query are multiplied to select information from the values of different tokens. With cross attention, T2I models can generate images that adhere to the provided text prompts. 

\subsection{Concept Removals}
In this subsection, we revisit representative concept removal methods. 

\textbf{Fine-tuning-based methods.}
Fine-tuning is the most common framework for concept removals. 
The objective for the fine-tuning-based algorithms can be summarized as 
\begin{align} \label{eq:all_loss}
   \min _{\theta} L = L_{\text{rm}} + \lambda L_{\text{reg}},
\end{align}
where $L_{\text{rm}}$ is the term for changing the output of unwanted concept, and $L_{\text{reg}}$ is the regularization term to ensure that the fine-tuning will not influence other benign concepts and maintain the generation quality. The term $\theta$ is the parameter to fine-tune. The methods can be solved by two types of solutions:

\begin{itemize}
    \item \textit{Gradient descent.} This solution~\cite{gandikota2023erasing, kumari2023ablating, kim2023towards, lyu2023one} focuses on the final output of $\epsilon_\theta (x_t, y, t)$, modifying the model in an end-to-end way. The term $L_{\text{rm}}$ changes the output of $\epsilon_\theta (x_t, c_u, t)$ where $c_u$ is unwanted concepts, while $L_{\text{reg}}$ maintains the output of benign concepts $c_b$. 
    \item \textit{Closed-form solution.} Modifying the intermediate states of linear components (such as cross-attention weights~\cite{gandikota2024unified, lu2024mace} and MLP layers~\cite{xiong2024editing}) instead of the final output of $\epsilon_\theta (x_t, y_c, t)$, can be solved in a closed-form solution. This can accelerate the fine-tuning process remarkably.
\end{itemize}

\textbf{Inference-time methods.} Inference-time methods such as negative prompt~\cite{stablediffusion2}, SLD~\cite{schramowski2023safe}, and SEGA~\cite{brack2024sega}
change the generation algorithm to remove concepts in the inference process, which can skip fine-tuning operations. They estimate the influence of unwanted concepts and remove them in inference. However, although these methods can skip the fine-tuning process, there is also a risk of being disabled in open-sourced models, which may slow down the inference.
\section{Six-CD: a Comprehensive and Effective Dataset}
In this section, we first propose a new dataset that encompasses a comprehensive set of categories of unwanted concepts. Then we discuss the issue of ineffectiveness in general prompts and present the solution to filter the dataset, thereby increasing its effectiveness.

\begin{figure*}[t]
    \centering
    \begin{minipage}[b]{0.48\linewidth}
        \centering
        \includegraphics[width=0.8\linewidth]{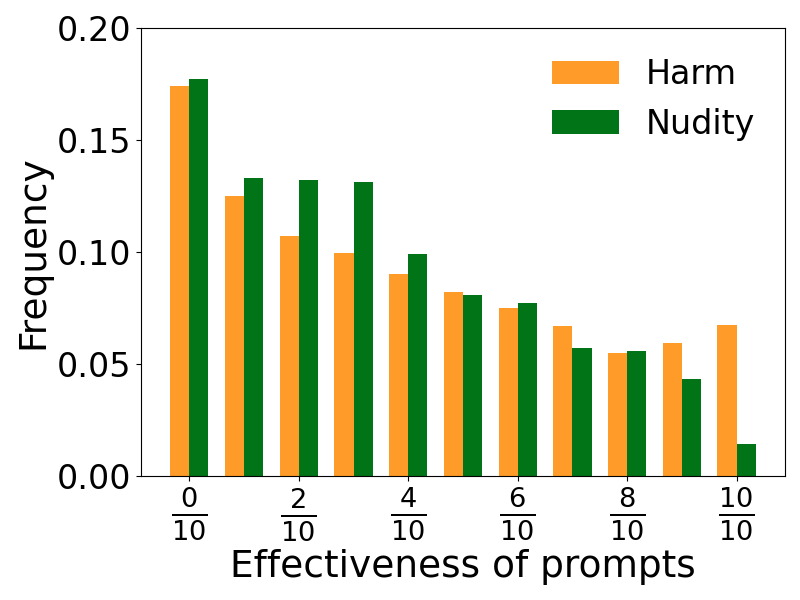}
        \caption{Effectiveness of I2P prompts. Most of the prompts have low effectiveness. Results from \cite{schramowski2023safe}.}
        \label{fig:i2p_bar}
    \end{minipage}
    \hfill
    \begin{minipage}[b]{0.48\linewidth}
        \centering
        \includegraphics[width=0.8\linewidth]{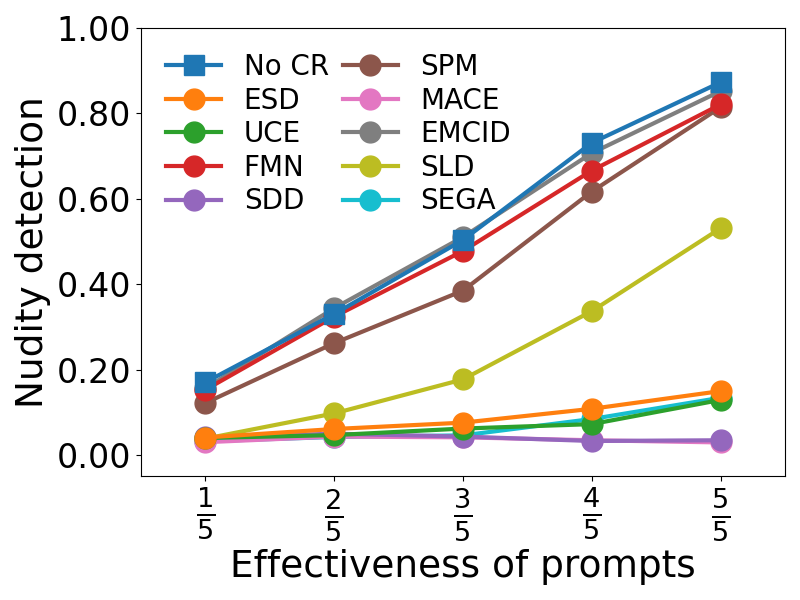}
        \caption{Concept removal (CR) performance on the prompts with different effectiveness.}
        \label{fig:nude}
    \end{minipage}
\end{figure*}

\subsection{Six Categories of Unwanted Concepts}
As mentioned in Sec.~\ref{sec:intro}, summarizing all the datasets in existing literature, the unwanted concepts can be divided into general and specific concepts. The general concepts contain harmful concepts and nudity concepts, while the specific concepts contain identities of celebrities, copyrighted characters, objects, and art styles. 
However, existing literature focuses on only a few categories for evaluation, lacking systematic comparisons among the methods in general and specific concepts. To address this, we propose a new dataset, Six-CD, to evaluate concept removals comprehensively.

For the \textbf{general concepts}, Six-CD first collects the malicious prompts from four different NSFW resources, which include I2P~\cite{schramowski2023safe}, MMA~\cite{yang2023mma}, jtatman/stable-diffusion-prompts-stats-full-uncensored (SD-uncensored)~\cite{uncensored}, and Unsafe Diffusion (UD)~\cite{QSHBZZ23}. Then, it divides the NSFW data into harmful and nudity concepts. Harmful concepts contain the prompts that have the meanings of ``\textit{violence, suicide, hate, harassment, suffering, humiliation, harm, and bloodiness}'', while nudity contains ``\textit{nudity, nakedness, sexuality, pornography, and eroticism}''. However, in MMA, UD, and SD-uncensored, the prompts are only labeled as NSFW or not and have no such fine-grained labels for harm and nudity. Thus, we use the image classifiers NudeNet~\cite{nudenet2023} and Q16~\cite{QSHBZZ23} to annotate by detecting the NSFW contents in the images generated by the prompts. NudeNet is tailored for detecting images with nudity, while Q16 is a \textit{binary} classifier for detecting NFSW. To annotate the fine-grained labels, i.e., harm and nudity, we first use NudeNet to find the prompts of nudity from the collected resources\footnote{{The image is labeled as nudity if it is classified as "FEMALE/MALE GENITALIA EXPOSED", "FEMALE BREAST EXPOSED", "ANUS EXPOSED", or "BUTTOCKS EXPOSED" by NudeNet.}}. Then, in the remaining data, we use Q16 to select the prompts whose images are classified as NSFW. Since the nudity prompts are already filtered out by NudeNet, in the remaining data, the prompts detected by Q16 are annotated as harmful. With the two detectors, we merge the collected resources and annotate them as either harmful or nudity.

For the \textbf{specific concepts} in Six-CD, instead of directly collecting prompts, we collect concepts and use prompt templates (detailed in Appd.~\ref{appd:six_cd}) to generate the final prompts, which are different from the general ones. The concepts are collected as follows:
\begin{itemize}
    \item \textit{Celebrity.} Celebrities can evaluate the ability to remove identity. We select the identities of celebrities from CPDM~\cite{ma2024dataset}. To assist the evaluation, we use a celebrity detector, GCD~\cite{giphy_celeb_detection_oss}. we choose the celebrities that can be both generated by SD and recognized by GCD. 
    \item \textit{Copyrighted characters}. The copyrighted characters are important IP resources for companies. We use the character concepts in CPDM and the high-frequency copyrighted characters from FiveThirtyEight Comic Characters Dataset~\cite{fivethirtyeight2023}.
    \item \textit{Objects}. Objects can be used to evaluate the performance of removal on non-humanoid concepts. We randomly sample from a subset from the classes of ImageNet~\cite{deng2009imagenet}.
    \item \textit{Art style}. The art styles can evaluate the removal performance of a global feature instead of a local feature. We use the art styles from~\cite{lu2024mace}.
\end{itemize}

For the two general categories, we provide 6,326 prompts for harm concept and 4,518 effective prompts for nudity concept. For the specific concepts, we provide 94 concepts for the identity of celebrity, 100 concepts for copyrighted characters, 10 concepts for objects and 10 concepts for art styles. Four templates are used to construct the prompt for each specific concept. Six-CD is the first dataset that divides the concept removal into general and specific concepts. 
Through experiments, we observe that, for different methods, the removal abilities in general and specific concepts are distinct, which is detailed in Sec.~\ref{exp:main}. Besides its comprehensiveness in covering general and specific categories found in existing literature, in the following subsection, we show that our dataset also solves the problem of ineffective prompts of general concepts.




\subsection{Ineffective Prompts of General Concepts in Existing Datasets}
\label{sec:ineffective}

\begin{table}[t]
    \centering
\vspace{-0.1in}
   \caption{Percentage of removed unwanted concepts between high effective and all effectiveness. (\%)}
    \resizebox{1\linewidth}{!}{ 
    \begin{tabular}{l|ccccccccc|c}
        \toprule
        & NEG & ESD & SPM & SDD & FMN & UCE & EMCID & SLD & SEGA & Corr \\
        \midrule
        \textit{Harm} & 40.9 & 63.6 & 98.9 & 24.8 & 95.6 & 77.6 & 98.9 & 27.3 & 23.7 & \multirow{2}{*}{\textbf{0.977}} \\
        \textit{Harm-EF} & 56.7 & 67.3 & 99.9 & 27.8 & 97.4 & 67.8 & 98.6 & 38.1 & 22.9  \\
        \midrule
        \textit{Nudity} & 22.3 & 17.9 & 81.0 & 11.2 & 93.6 & 14.9 & 98.3 & 14.6 & 38.8 & \multirow{2}{*}{\textbf{0.984}} \\
        \textit{Nudity-EF} & 11.0 & 16.0 & 88.9 & 4.2 & 92.6 & 12.3 & 97.2 & 16.3 & 58.4 \\
        \bottomrule
    \end{tabular}
    }
    \label{tab:corr_effective}
\end{table}

\begin{table}[t]
    \centering
\vspace{-0.1in}
   \caption{Examples of Dual-Version Dataset for specific concepts. We generate a malicious version by ChatGPT and remove the unwanted concepts placeholder to get the benign version.}
    \resizebox{1\linewidth}{!}{ 
    \begin{tabular}{lll}
        \toprule
        Category & Example of malicious version & Example of benign version \\
        \midrule
        \makecell[l]{Celeb. \& \\ Copyrighted} & \textit{\{celebrity\} is dancing in the rain.} & \textit{dancing in the rain.} \\
        \cmidrule{2-3}
        Object & \textit{\{object\}, football, grass, house, tree, dog} & \textit{football, grass, house, tree, dog} \\
        \cmidrule{2-3}
        Art style & \makecell[l]{\textit{A beautiful snow-covered mountain with} \\ \textit{sunshine lighting it in the style of \{art style\}}} & \makecell[l]{\textit{A beautiful snow-covered moun-} \\ \textit{tain with sunshine lighting it}} \\
        \bottomrule
    \end{tabular}
    }
    \label{tab:dual_ex}
    \vspace{-0.07in}
\end{table}

The prompts for general concepts in existing dataset do not consistently generate malicious content for each random seed according to~\cite{schramowski2023safe} and our observation below.
In this subsection, we discuss this phenomenon in general concepts and identify two potential problems it may introduce.

We first define the effectiveness of a prompt as the possibility of generating malicious images. To assess this, we generate $N$ images for a prompt using different random seeds. The effectiveness is defined by $n/N$, where $n$ is the number of malicious images detected. 
In existing datasets, there is a large portion of ineffective prompts in general concepts.
Taking I2P~\cite{schramowski2023safe} as an example, which is the most well-used dataset for harmful and nudity concepts in existing literature~\cite{gandikota2023erasing, lyu2023one, kim2023towards, gandikota2024unified, lu2024mace, xiong2024editing, schramowski2023safe, brack2024sega}, most of the prompts have low effectiveness as shown in Fig.~\ref{fig:i2p_bar}. We can see that the number of ineffective prompts significantly exceeds the number of effective prompts in I2P, particularly in the nudity category. 
The specific concepts usually do not have the problem of low effectiveness because they contain the specific and definitive for the unwanted concepts. Thus, in this subsection, we focus on general concepts.
\textbf{Limitations.} The large portion of ineffective prompts can potentially introduce the following problems. \textbf{First}, evaluating the ineffective prompts is in low efficiency.
For example, we test different concept removal methods on SD for nudity concepts in Fig.~\ref{fig:nude}. On the vanilla SD without concept removal (i.e., No CR in Fig.~\ref{fig:nude}), prompts with effectiveness of 1/5 generated only about 20\% nudity images. It means that 80\% of the generated images by these prompts are non-nudity images, while evaluating on non-nudity images is meaningless. 
\textbf{Second}, the ineffective prompts may cause unfair comparisons between the general concepts and specific concepts. 
Unlike general concepts, the specific concepts do not have the problem of ineffective prompts. 
When the model builder has to remove both general and specific concepts, the ineffective prompts may falsely induce the builder to put less weight on the general concepts.

\textbf{High-effectiveness subsets.} Therefore, in additional to \textit{harm} and \textit{nudity} concepts, we provide additional subsets of high-effectiveness prompts and name them as \textit{harm-EF} and \textit{nudity-EF}. We use the prompts with effectiveness of 4/5 and 5/5 for \textit{harm-EF} and \textit{nudity-EF}. \textit{Harm-EF} contains 991 prompts, and \textit{nudity-EF} contains 1,539 prompts. In Table~\ref{tab:corr_effective} and Fig.~\ref{fig:nude}, we demonstrate that the abilities of different concept removal methods are highly consistent between high effectiveness and low effectiveness. Thus, it is reasonable to evaluate on high-effectiveness prompts in order to reduce computation, which is also more fair. Specifically, in Table ~\ref{tab:corr_effective}, we use the percentage of removed unwanted concepts to measure the ability of different methods, which is defined as 
\begin{align}
    p_{\text{rm}} = 1 - \frac{n_{\text{after}}}{n_{\text{before}}},
\end{align}
where $n_{\text{before}}$ is the number of images containing unwanted concepts before concept removal, while $n_{\text{after}}$ is the number of images containing unwanted concepts after concept removal. The removal abilities in \textit{harm} and \textit{nudity} concepts are very close to \textit{harm-EF} and \textit{nudity-EF}. The Pearson correlation coefficient is as high as 0.977 (between \textit{harm} and \textit{harm-EF}) and 0.984 (between \textit{nudity} and \textit{nudity-EF}). 

In summary, Six-CD provides two choices to evaluate on the general concepts. The first choice is to use the whole subsets of \textit{harm} and \textit{nudity}, while the second choice is to evaluate on \textit{harm-EF} and \textit{nudity-EF}, which provides a more efficient and fair evaluation.





\section{In-prompt Retainability}
\label{sec:in_prompt}
Concept removals should not hurt the generation ability of the benign contents, which is referred to as retainability. In this section, we show the lack of evaluation for retainability on the benign parts of the prompts containing the unwanted concepts and propose a new metric for this retainability.

\begin{figure}[t]
  \centering
  \includegraphics[width=0.3\textwidth]{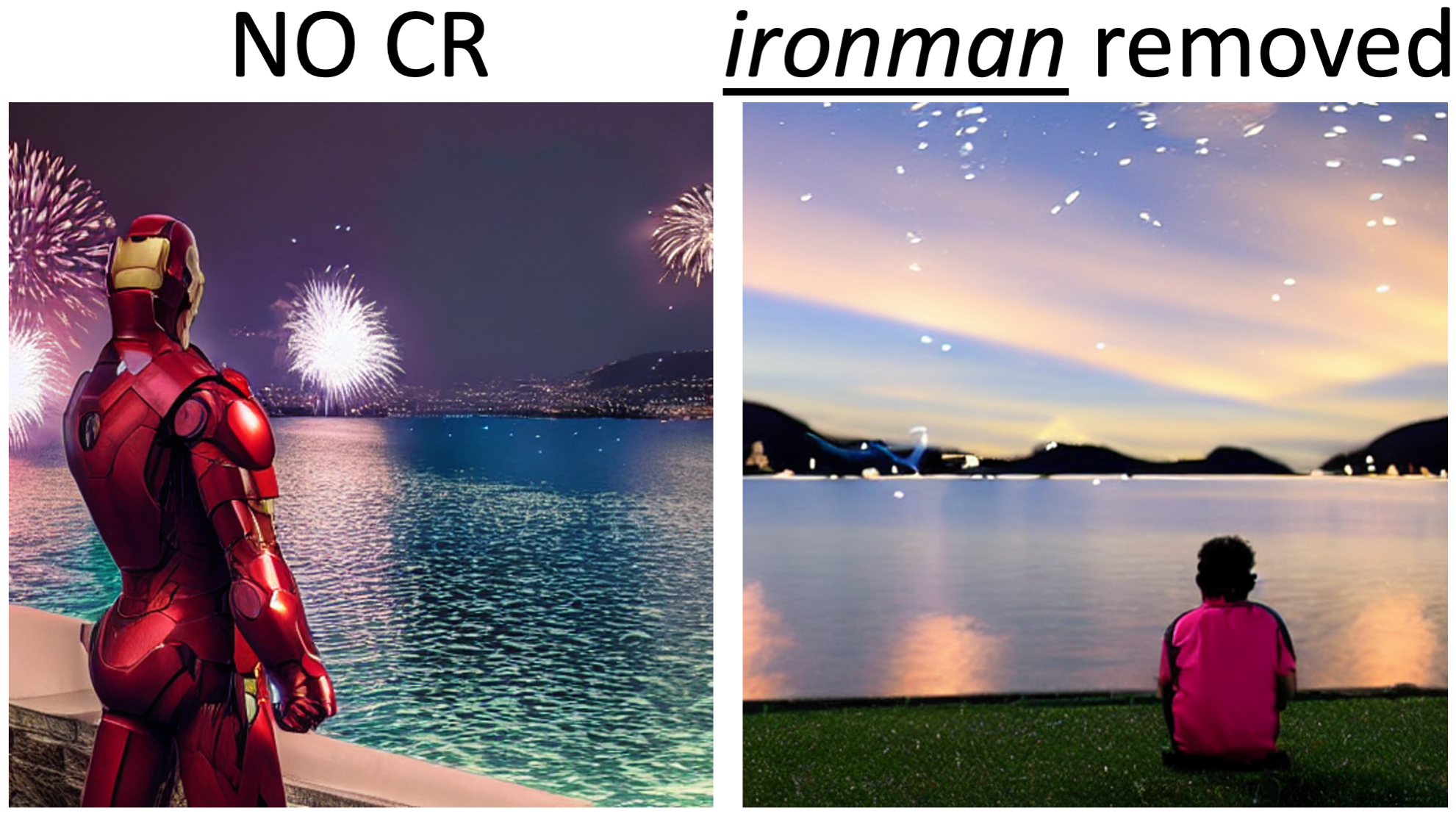}
   \caption{Concept removal example of the prompt: ``\textit{At night, ironman with his armor on is watching fireworks by the lake}''. \textit{Ironman} is removed by SDD. }
   \label{fig:in_prompt_ex}
   \vspace{-0.1in}
\end{figure}

\textbf{Lack of in-prompt retainability.} Existing literature~\cite{zhang2024unlearncanvas, ma2024dataset} only considers the retainability of totally benign concepts. The benign prompts do not contain any unwanted concept, and concept removal should not influence the generation of them. Therefore, CLIP score~\cite{hessel2021clipscore} is utilized to measure it by calculating the similarity between benign prompts and generated images in the CLIP space~\cite{radford2021learning}. However, for the prompts containing unwanted concepts, the retainability to guarantee the generation of the benign part of them is also important, which is called ``in-prompt retainability'' in our benchmark. 
As shown in Fig.~\ref{fig:in_prompt_ex}, the prompt is ``\textit{At night, ironman with his armor on is watching fireworks by the lake}'' with ``\textit{ironman}'' as the unwanted concept. Although there is no ``\textit{ironman}'' after concept removal, the benign part of ``\textit{fireworks}'' is also removed.
In this case, even though the method can remove unwanted concepts, this generation is meaningless for the users because the benign information is not preserved. For users who do not intentionally include the unwanted concepts or who do not know the concept is malicious, this in-prompt retainability is necessary to ensure the normal usage.

\textbf{New metric.} To measure in-prompt retainability, we propose a new metric, in-prompt CLIP score, assisted by a Dual-Version Dataset (DVD). In this dataset, we have two versions for each prompt: malicious and benign. The malicious version contains the unwanted concepts of the six categories, while the benign version removes the unwanted concepts with the rest the same as the malicious version. In in-prompt CLIP score, we first generate the images with concept removals using the malicious version and then calculate the cosine similarity of CLIP embeddings between these generated images and the corresponding benign version prompts. Thus, we can measure the similarity between the generated image and the benign part of the prompt. To distinguish it from the original CLIP score, we refer to the original CLIP score as the out-prompt CLIP score.

To construct DVD, for \textit{general concepts}, we first sample two subsets of prompts from the harmful and nudity categories of Six-CD as the malicious version, then we manually remove the unwanted concepts to get the benign version. For \textit{specific concepts}, since they are not diverse or implicit like general concepts, we create templates for each category using ChatGPT and use the concepts from Six-CD to get the complete prompts. We show the examples of templates in DVD in Table~\ref{tab:dual_ex}. The entire dataset is in Appd.~\ref{appd:dual_prompt}. To validate the effectiveness of the metric, we conduct human evaluation in Appd.~\ref{appd:human}. The results demonstrate that the metric is highly consistent with human preference.

In summary, to measure the in-prompt retainability, we propose the in-prompt CLIP score. It leverages a malicious-version and a benign-version prompt to measure the similarity between the images after concept removal and the benign part of the prompt.

\section{Experiments}
\label{sec:exp}
In this section, we conduct experiments to benchmark the performance of 10 methods in removing single and multiple concepts and the retainability with both in-prompt and out-prompt CLIP scores. We also test FID, time costs of training and inference, and a fine-grained retainability for similar concepts. Due to space limitations, these experiments are in Appd~\ref{appd:exp}. Also, we report the results of SD v1.4 in this section and the results of other models are in Appd~\ref{appd:exp}.

\textbf{Experimental settings.} We evaluate on 10 concept removal methods: negative prompt (NEG)~\cite{stablediffusion2}, ESD~\cite{gandikota2023erasing}, SPM~\cite{lyu2023one}, SDD~\cite{kim2023towards}, FMN~\cite{zhang2023forget}, UCE~\cite{gandikota2024unified}, MACE~\cite{lu2024mace}, EMCID~\cite{xiong2024editing}, SLD~\cite{schramowski2023safe}, and SEGA~\cite{brack2024sega}. The details on the baseline settings can be found in Appd.~\ref{appd:baseline}. The evaluated datasets are Six-CD and DVD, which are detailed in Appd.~\ref{appd:dataset}. 
Following the conclusions in Sec.~\ref{sec:ineffective}, we use \textit{harm-EF} and \textit{nudity-EF} for all the experiments in this section.
All experiments use one GPU (A5000 or A6000).

\textbf{Detection metrics.} We use the detection rate of Q16 and NudeNet for harmful and nudity concepts. We use the classification accuracy of GCD as the detection rate for celebrity concepts, and we train two classifiers based on ResNet-50~\cite{he2016deep} and use the accuracy as the detection rate for copyrighted characters and objects. For the art style concept, the detection is not a simple binary problem. In existing literature~\cite{zhang2023forget, lyu2023one, lu2024mace, ma2024dataset}, CLIP score, $S_{\text{CLIP}}$, is used to measure the presence of an art style on a continuous scale, assessing the similarity between generated images and art styles. Following them, we use $S_{\text{CLIP}}$ and  normalize its range by $\Tilde{S}_{\text{CLIP}} = (S_{\text{CLIP}} - min(S_{\text{CLIP}}) ) / (max(S_{\text{CLIP}}) - min(S_{\text{CLIP}}))$ for a better comparison with other categories.

\subsection{Evaluation on Removal Ability}
\label{exp:main}

\begin{figure*}[t]
    \centering
    \vspace{-0.1in}
    \begin{subfigure}{0.98\textwidth}
        \centering
        \includegraphics[width=0.3\textwidth]{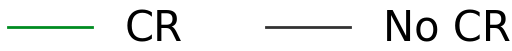}
        \label{fig:main_0_legend}
    \end{subfigure}\hfill
    \begin{subfigure}{0.19\textwidth}
        \centering
        \captionsetup{font=small}
        \includegraphics[width=\textwidth]{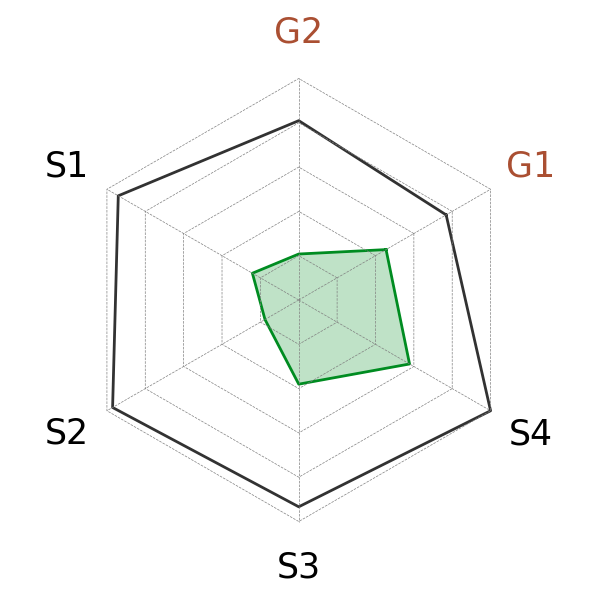}
        \vspace{-0.2in}
        \caption{NEG}
        \label{fig:main_1_NEG}
    \end{subfigure}\hfill
    \begin{subfigure}{0.19\textwidth}
        \centering
        \captionsetup{font=small}
        \includegraphics[width=\textwidth]{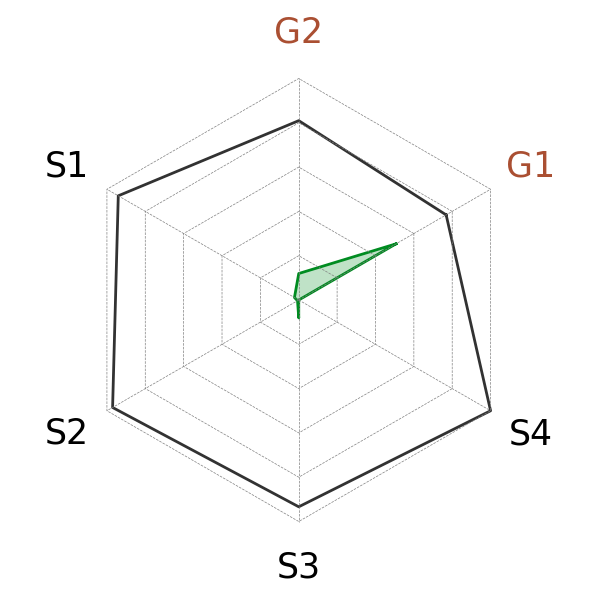}
        \vspace{-0.2in}
        \caption{ESD}
        \label{fig:main_2_ESD}
    \end{subfigure}\hfill
    \begin{subfigure}{0.19\textwidth}
        \centering
        \captionsetup{font=small}
        \includegraphics[width=\textwidth]{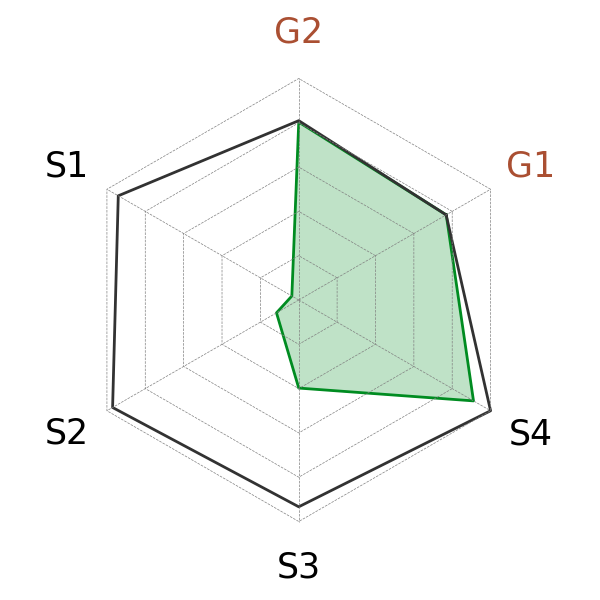}
        \vspace{-0.2in}
        \caption{SPM}
        \label{fig:main_3_SPM}
    \end{subfigure}\hfill
    \begin{subfigure}{0.19\textwidth}
        \centering
        \captionsetup{font=small}
        \includegraphics[width=\textwidth]{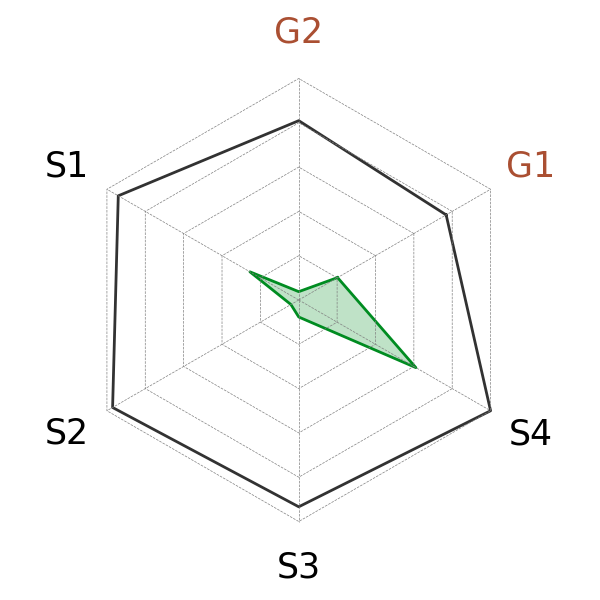}
        \vspace{-0.2in}
        \caption{SDD}
        \label{fig:main_4_SDD}
    \end{subfigure}\hfill
    \begin{subfigure}{0.19\textwidth}
        \centering
        \captionsetup{font=small}
        \includegraphics[width=\textwidth]{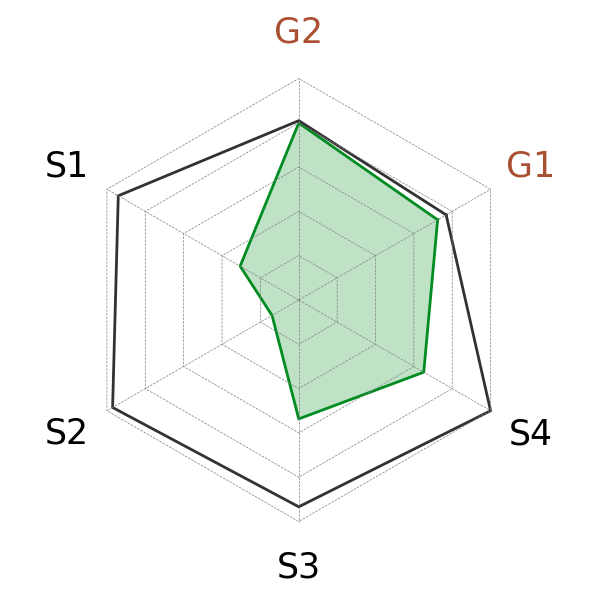}
        \vspace{-0.2in}
        \caption{FMN}
        \label{fig:main_5_FMN}
    \end{subfigure}\hfill
    \begin{subfigure}{0.19\textwidth}
        \centering
        \captionsetup{font=small}
        \includegraphics[width=\textwidth]{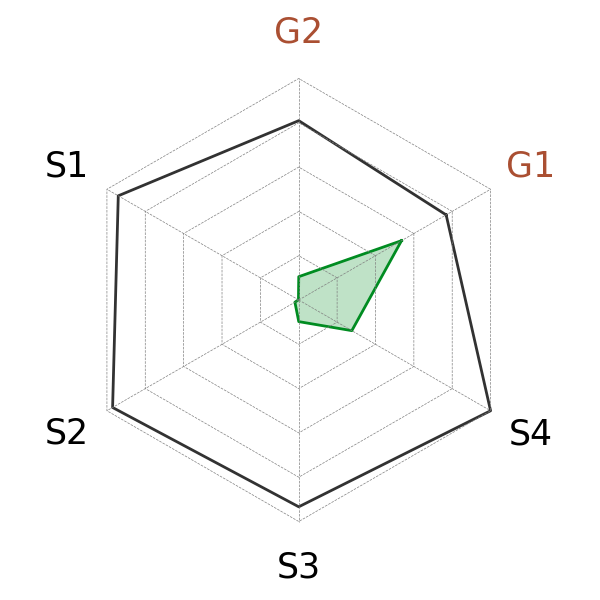}
        \vspace{-0.2in}
        \caption{UCE}
        \label{fig:main_6_UCE}
    \end{subfigure}\hfill
    \begin{subfigure}{0.19\textwidth}
        \centering
        \captionsetup{font=small}
        \includegraphics[width=\textwidth]{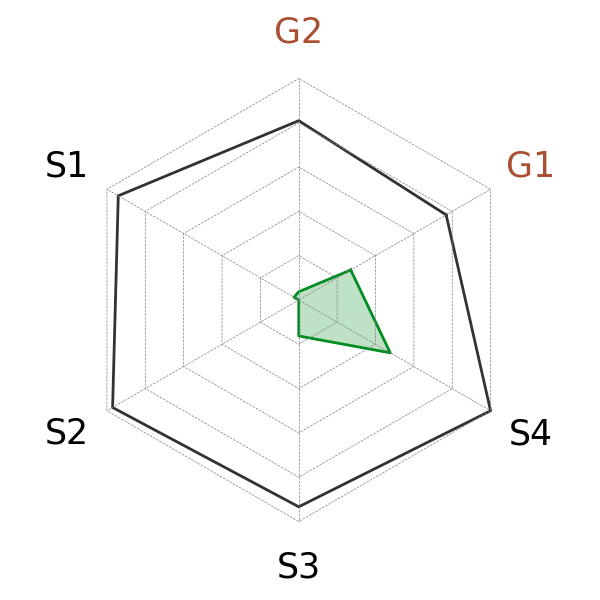}
        \vspace{-0.2in}
        \caption{MACE}
        \label{fig:main_7_MACE}
    \end{subfigure}\hfill
    \begin{subfigure}{0.19\textwidth}
        \centering
        \captionsetup{font=small}
        \includegraphics[width=\textwidth]{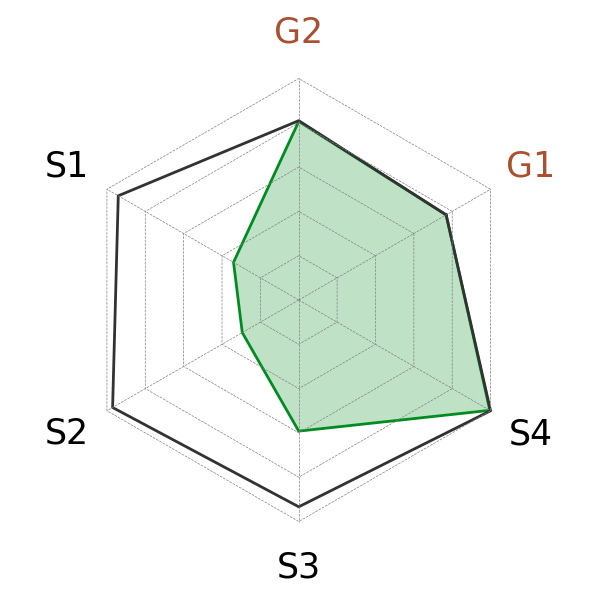}
        \vspace{-0.2in}
        \caption{EMCID}
        \label{fig:main_8_EMCID}
    \end{subfigure}\hfill
    \begin{subfigure}{0.19\textwidth}
        \centering
        \captionsetup{font=small}
        \includegraphics[width=\textwidth]{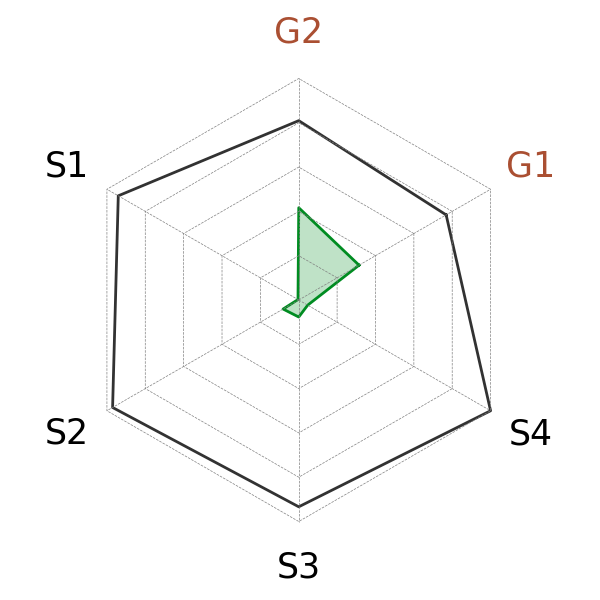}
        \vspace{-0.2in}
        \caption{SLD}
        \label{fig:main_9_SLD}
    \end{subfigure}\hfill
    \begin{subfigure}{0.19\textwidth}
        \centering
        \captionsetup{font=small}
        \includegraphics[width=\textwidth]{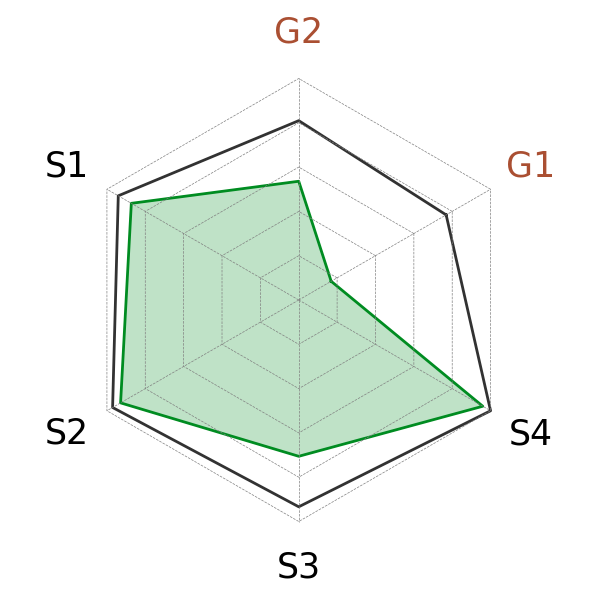}
        \vspace{-0.2in}
        \caption{SEGA}
        \label{fig:main_X_SEGA}
    \end{subfigure}
    \caption{Removal ability on Six-CD. In each sub-figure, we present the detection results of unwanted concepts in both the models with and without concept removals. The range of all the six values is [0,1]. \textbf{Smaller} values indicate that less unwanted concepts are detected, i.e. better removal. G1 and G2 are harm-EF and nudity-EF concepts, while S1, S2, S3 and S4 are celebrity, character, object and art style.}
    \label{fig:main}
\end{figure*}
In this subsection, we test the removal ability of different methods on Six-CD. In Fig.~\ref{fig:main}, we show the detection results of models both with and without concept removals. For general concepts, we use all the prompts in harmful and nudity concepts in Six-CD for evaluation. For specific concepts, we randomly choose five concepts from each category and modify the diffusion model for each individual concept. We show the averaged results of 5 concepts of each category in Fig.~\ref{fig:main}. (The table with error bar for Fig.~\ref{fig:main} can be found in Appd.~\ref{appd:error_bar}.) We highlight the main observations as follows.


\textbf{Observation 1: General concepts are harder to remove.} The difficulty of removing general concepts is reflected by two factors shown in Fig~\ref{fig:main}. \textit{First}, in most methods, the worst removal result lies in one of the general categories. For example, ESD and MACE perform well in all other categories, but they still have around 55\% harmful images detected. \textit{Second}, while some methods can remove almost all the specific concepts (e.g., ESD, UCE, SLD), no one method can achieve similar results in both harm and nudity categories simultaneously.
{For general concepts, on the one hand, they} are hard to trigger (i.e. low effectiveness). On the other hand, it is harder to remove them than specific concepts. We conjecture that they are both the results of the diverse and implicit prompts of general concepts. Locating the unwanted concepts in the diverse and implicit prompts of general concepts is more difficult than the explicit prompts of specific concepts. Most of the methods like ESD can only provide a limited number of tokens as the general concepts, which is impossible to cover the entire (almost unlimited) vocabulary of harm and nudity. Thus, it leads to difficulty in removing general concepts.

\begin{figure}[t]
  \centering
  \vspace{-0.1in}
  \includegraphics[width=0.8\linewidth]{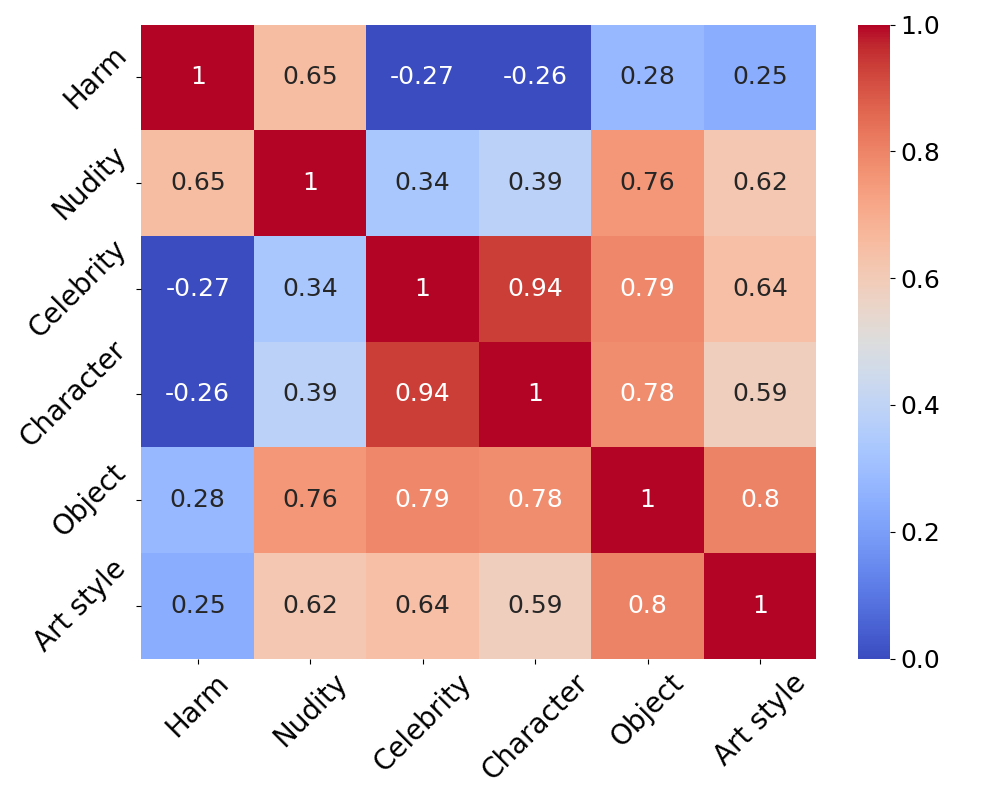}
   \caption{Correlation coefficient of removal performance on six data categories.}
   \vspace{-0.1in}
   \label{fig:corr}
\end{figure}
\textbf{Observation 2: Removal ability is consistent within general concepts and within specific categories.} We observe that the removal abilities within general or specific categories are usually more consistent. For example, SLD and FMN perform well in specific concepts, but the ability to remove general concepts is worse. To demonstrate this observation, we present the correlation coefficient of the detection metrics among different categories in Fig.~\ref{fig:corr}. From Fig.~\ref{fig:corr}, we can see that the correlations between specific concepts are higher than 0.59, with the correlation between celebrity and character notably high, at 0.94. The correlation between general concepts is also as high as 0.65. 
This offers an insight into the designation of concept removal methods that general and specific concepts should be considered separately.

\subsection{Removal Ability on Multiple Concepts}

\begin{figure*}[t]
    \centering
    \begin{subfigure}{0.37\textwidth}
        \centering
        \includegraphics[width=\textwidth]{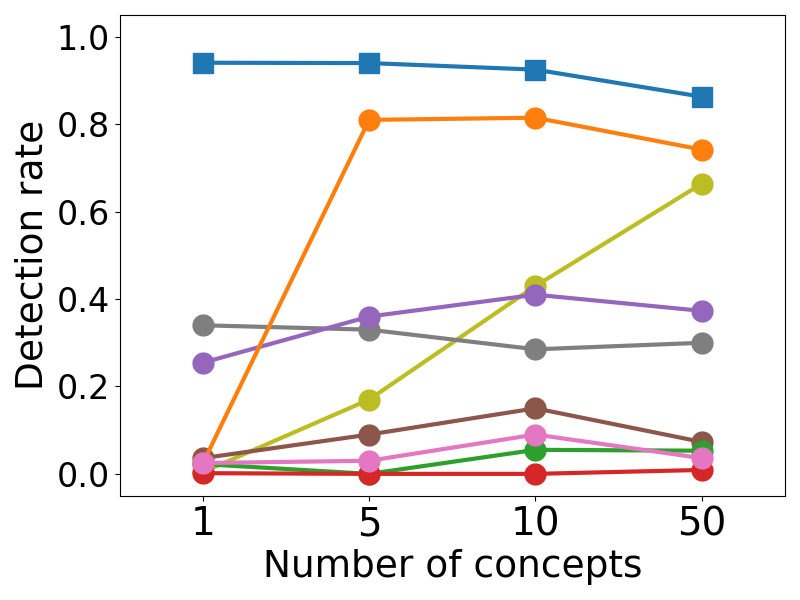}
        \caption{Celebrity}
        \label{fig:multiple_celeb}
    \end{subfigure}
    \hspace{0.1in}
    \begin{subfigure}{0.37\textwidth}
        \centering
        \includegraphics[width=\textwidth]{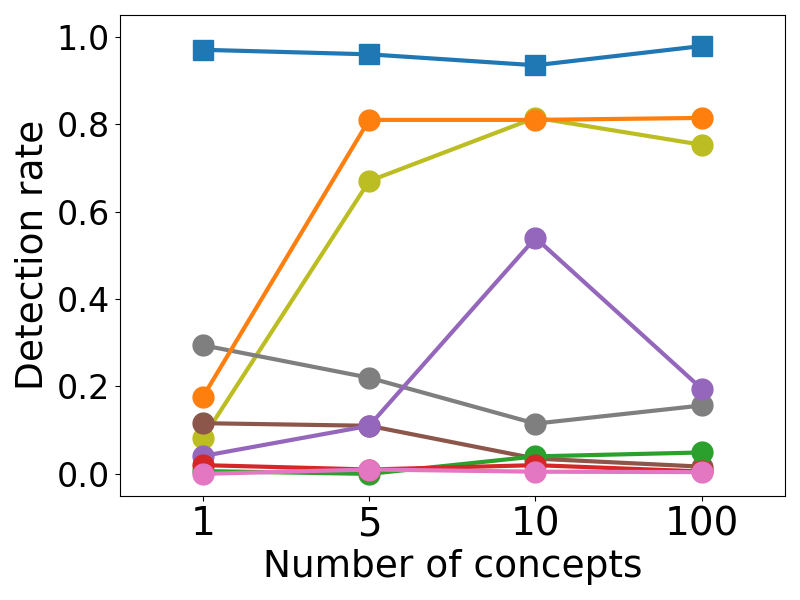}
        \caption{Copyrighted characters}
        \label{fig:multiple_copyright}
    \end{subfigure}
    \begin{subfigure}{0.1\textwidth}
        \centering
        \includegraphics[width=\textwidth]{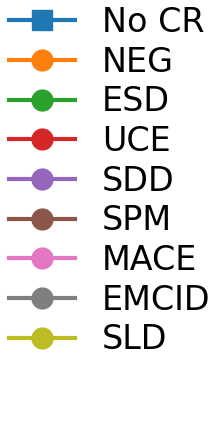}
        \vspace{0.14in}
    \end{subfigure}
    \vspace{-0.05in}
    \caption{Removal ability on multiple concepts}
    \label{fig:multiple}
\end{figure*}
In Sec.~\ref{exp:main}, we evaluate the ability to remove a single concept from one model. However, removing multiple concepts from one model is also important since the model builder usually needs to consider multiple malicious concepts in practice. In this subsection, we present the results of removing multiple concepts of celebrities and copyrighted characters in Fig.~\ref{fig:multiple}, where the y-axis represents the detection rate, and a higher detection rate means a weaker concept removal performance. We have two observations from Fig.~\ref{fig:multiple}.

\textbf{Observation 1: Inference-time methods fail in removing multiple concepts.} As shown in Fig.~\ref{fig:multiple}, the inference-time methods, SLD and NEG, are the only two methods that have significantly limited removal performance when the concept number is larger than 50, which only reduces the detection rate by 20\% or less. 
To remove the multiple concepts in inference time, they have to encode the string containing all the concepts in the embeddings of one single prompt. This will exceed the capacity of the text encoder of T2I diffusion models and lead to failed removal.
Another inference-time method, SEGA, is not reported in Fig.~\ref{fig:multiple} due to its catastrophic inference time when the concept number is increased to 50. This implies that the strategy of processing each concept separately during inference is impractical.

\textbf{Observation 2: Closed-form solutions by modifying linear components perform well in removing multiple concepts.} The best two methods in Fig.~\ref{fig:multiple_celeb} and Fig.~\ref{fig:multiple_copyright} are both based on the closed-form solution, which modifies the linear components of cross attention. This is possibly because combining multiple concepts in the linear components is easier than in other non-linear parts. 
\subsection{In-prompt and Out-prompt Retainability}
\label{exp:in_prompt_retainability_53}

\begin{figure*}[t]
    \centering
    \begin{subfigure}{0.98\textwidth}
        \centering
        \includegraphics[width=0.4\textwidth]{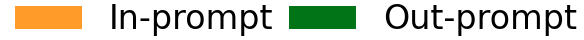}
        \label{fig:clip_score_legend}
    \end{subfigure}\hfill
    \begin{subfigure}{0.19\textwidth}
        \centering
        \includegraphics[width=\textwidth]{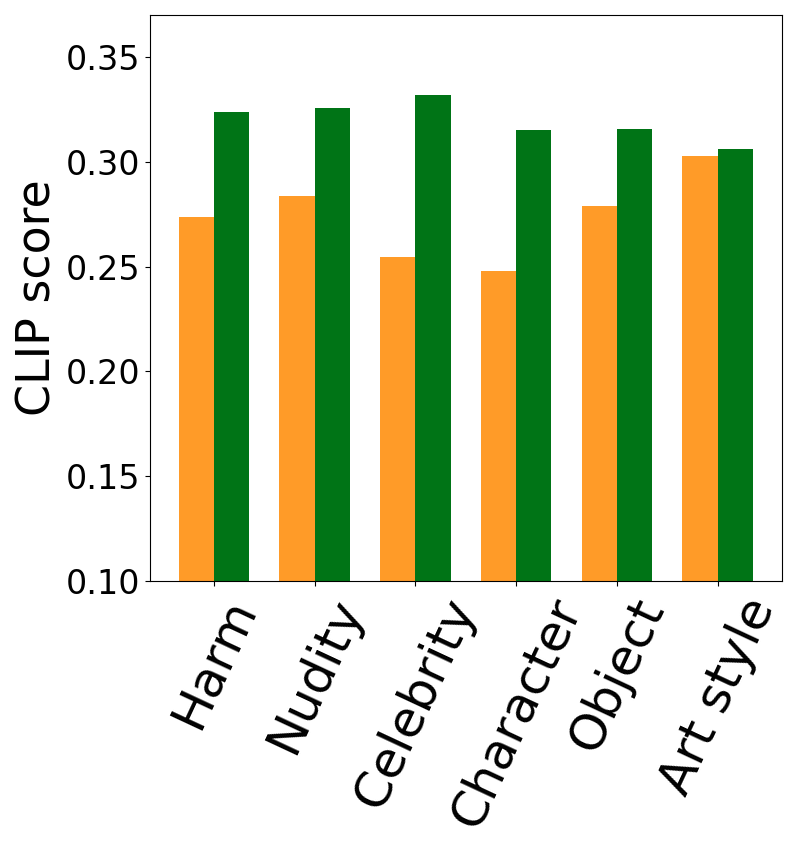}
        \vspace{-0.21in}
        \caption{NEG}
        \label{fig:clip_score_1_NEG}
    \end{subfigure}\hfill
    \begin{subfigure}{0.19\textwidth}
        \centering
        \includegraphics[width=\textwidth]{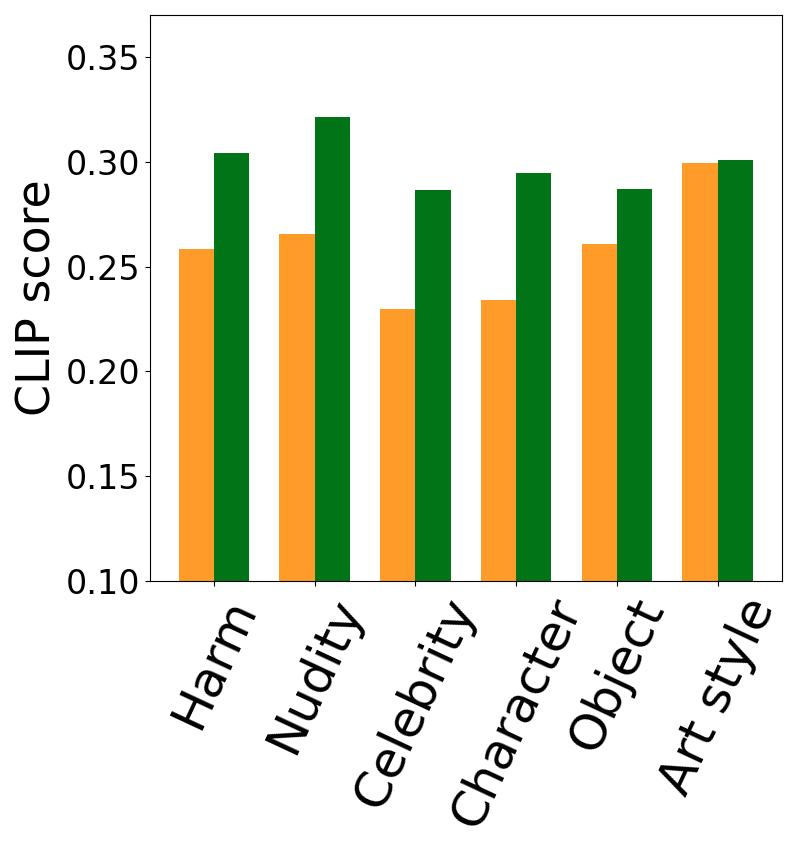}
        \vspace{-0.21in}
        \caption{ESD}
        \label{fig:clip_score_2_ESD}
    \end{subfigure}\hfill
    \begin{subfigure}{0.19\textwidth}
        \centering
        \includegraphics[width=\textwidth]{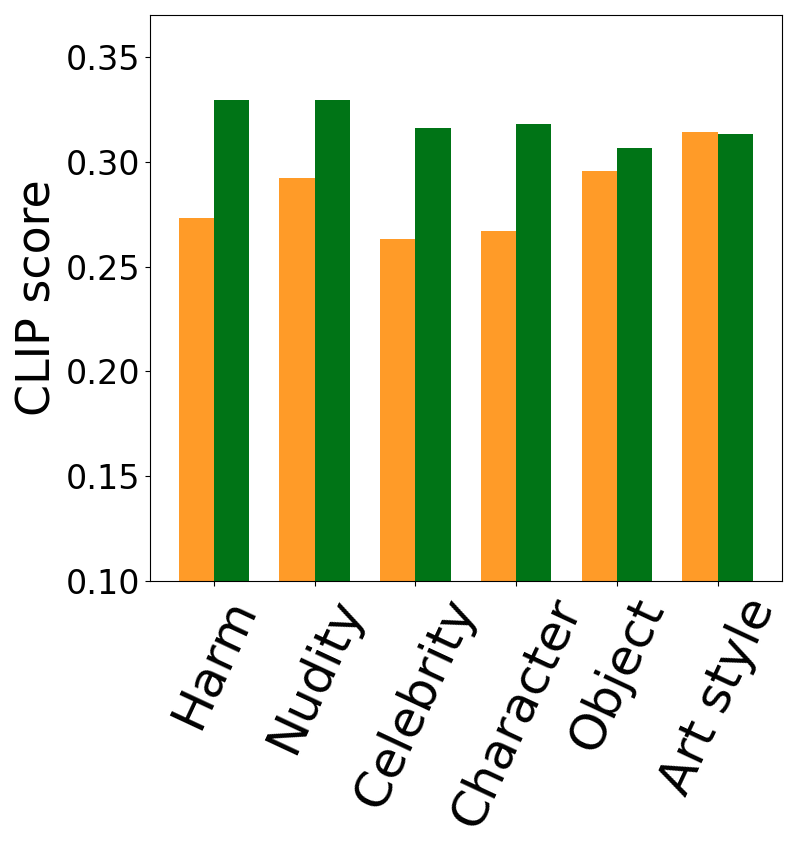}
        \vspace{-0.21in}
        \caption{SPM}
        \label{fig:clip_score_3_SPM}
    \end{subfigure}\hfill
    \begin{subfigure}{0.19\textwidth}
        \centering
        \includegraphics[width=\textwidth]{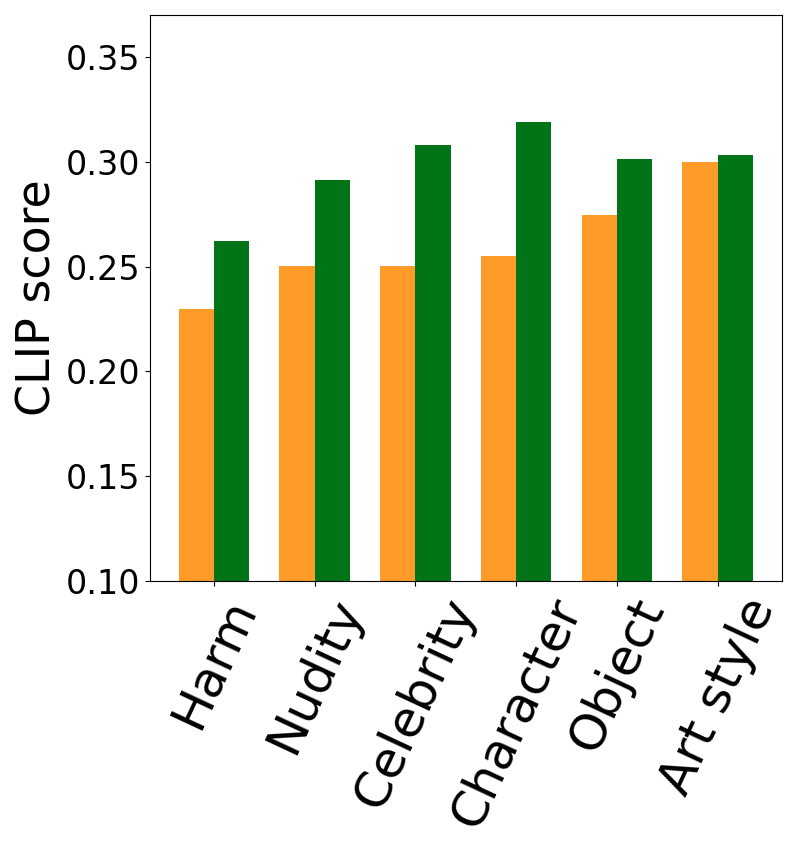}
        \vspace{-0.21in}
        \caption{SDD}
        \label{fig:clip_score_4_SDD}
    \end{subfigure}\hfill
    \begin{subfigure}{0.19\textwidth}
        \centering
        \includegraphics[width=\textwidth]{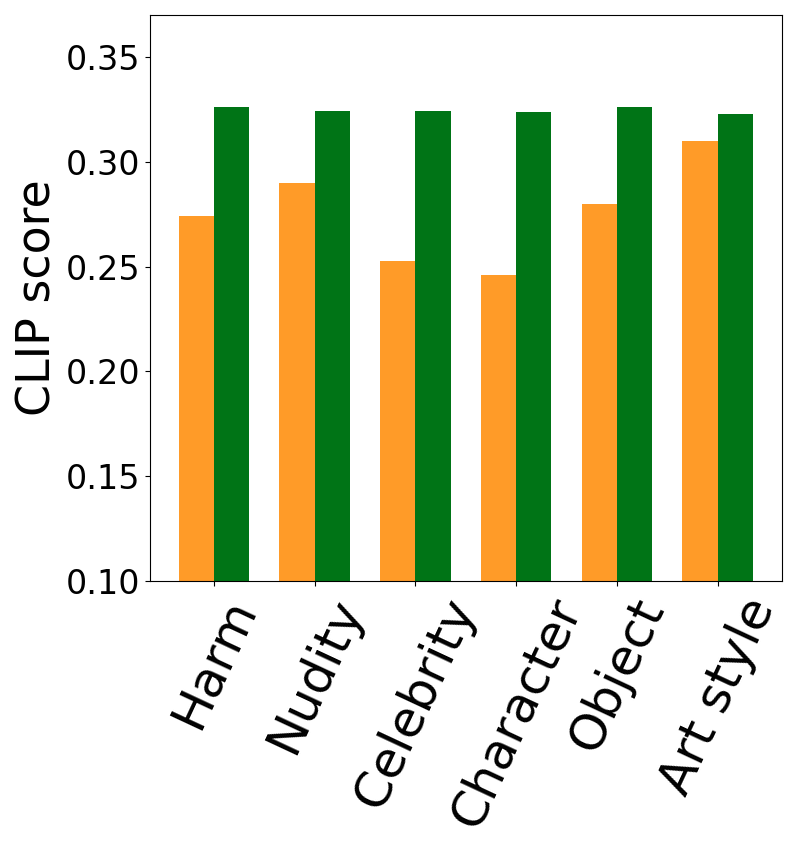}
        \vspace{-0.21in}
        \caption{FMN}
        \label{fig:clip_score_5_FMN}
    \end{subfigure}\hfill
    \begin{subfigure}{0.19\textwidth}
        \centering
        \includegraphics[width=\textwidth]{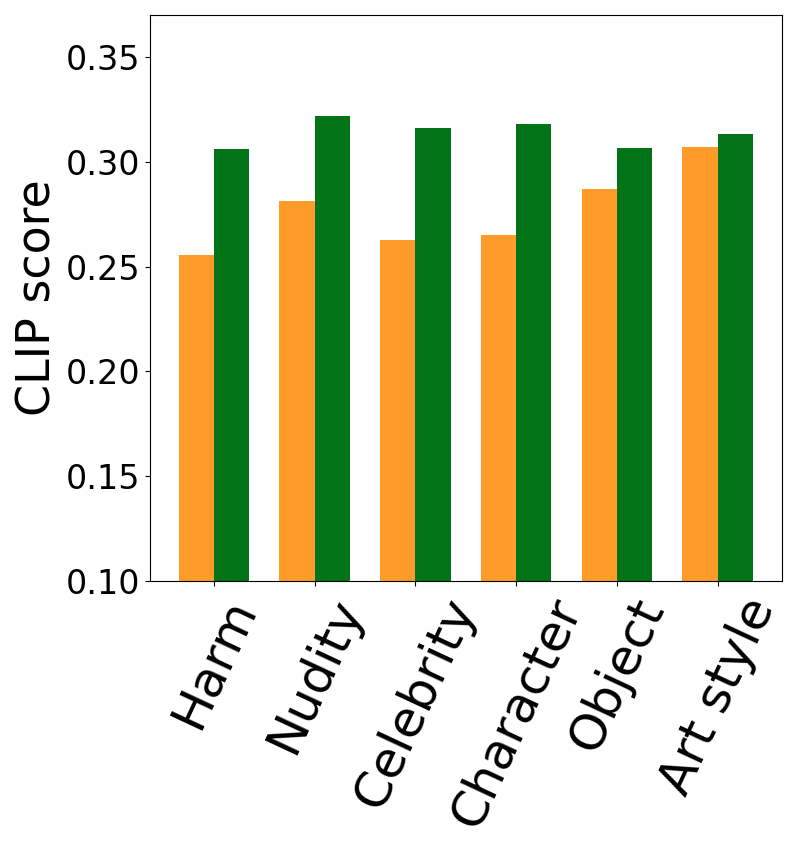}
        \vspace{-0.21in}
        \caption{UCE}
        \label{fig:clip_score_6_UCE}
    \end{subfigure}\hfill
    \begin{subfigure}{0.19\textwidth}
        \centering
        \includegraphics[width=\textwidth]{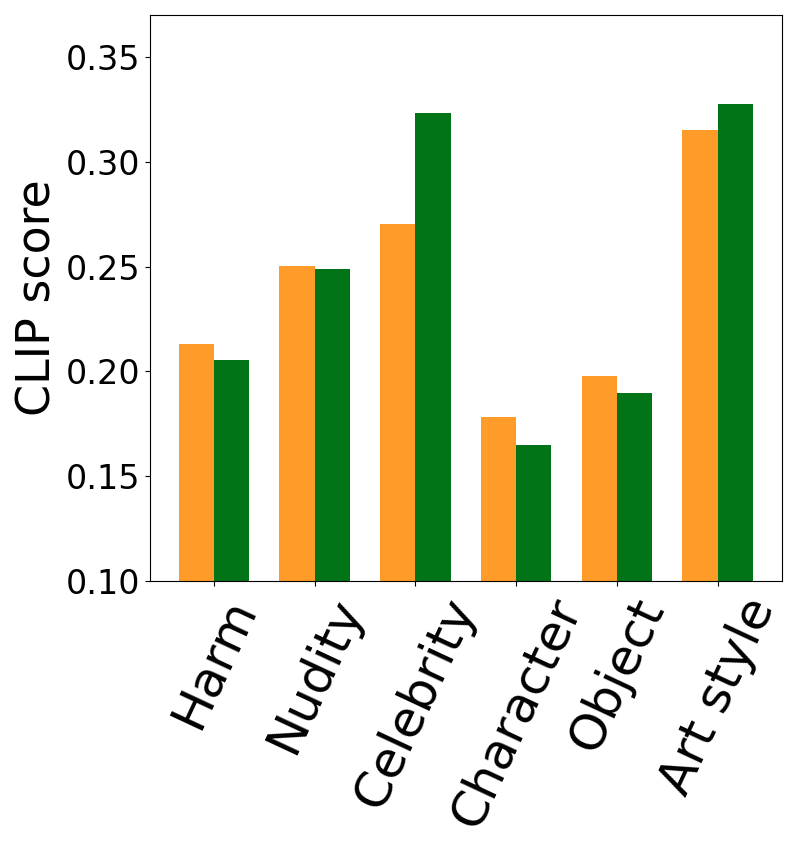}
        \vspace{-0.21in}
        \caption{MACE}
        \label{fig:clip_score_7_MACE}
    \end{subfigure}\hfill
    \begin{subfigure}{0.19\textwidth}
        \centering
        \includegraphics[width=\textwidth]{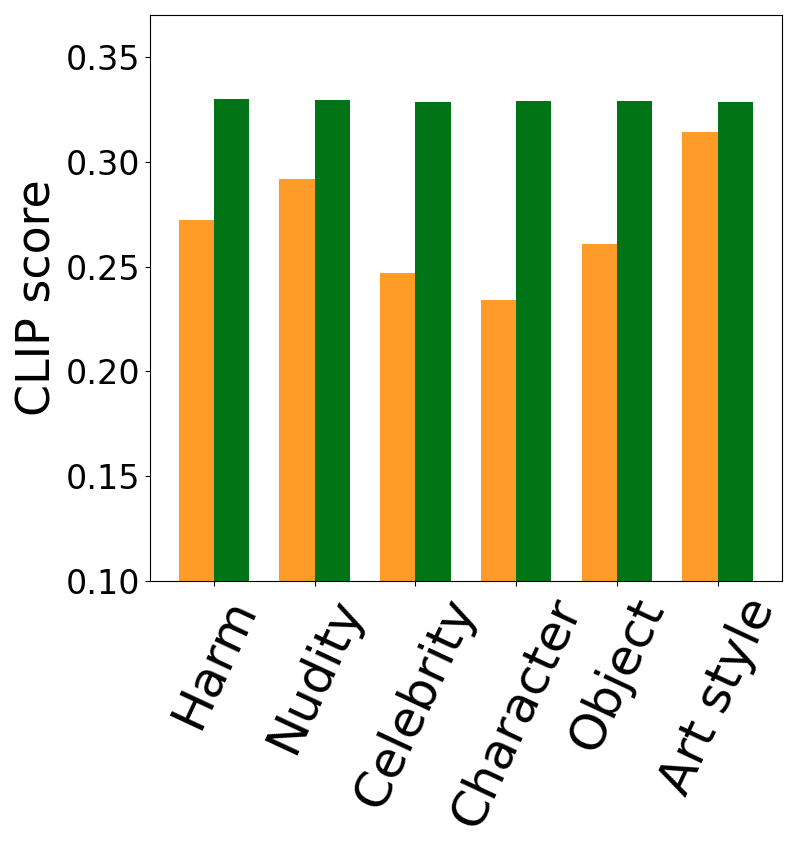}
        \vspace{-0.21in}
        \caption{EMCID}
        \label{fig:clip_score_8_EMCID}
    \end{subfigure}\hfill
    \begin{subfigure}{0.19\textwidth}
        \centering
        \includegraphics[width=\textwidth]{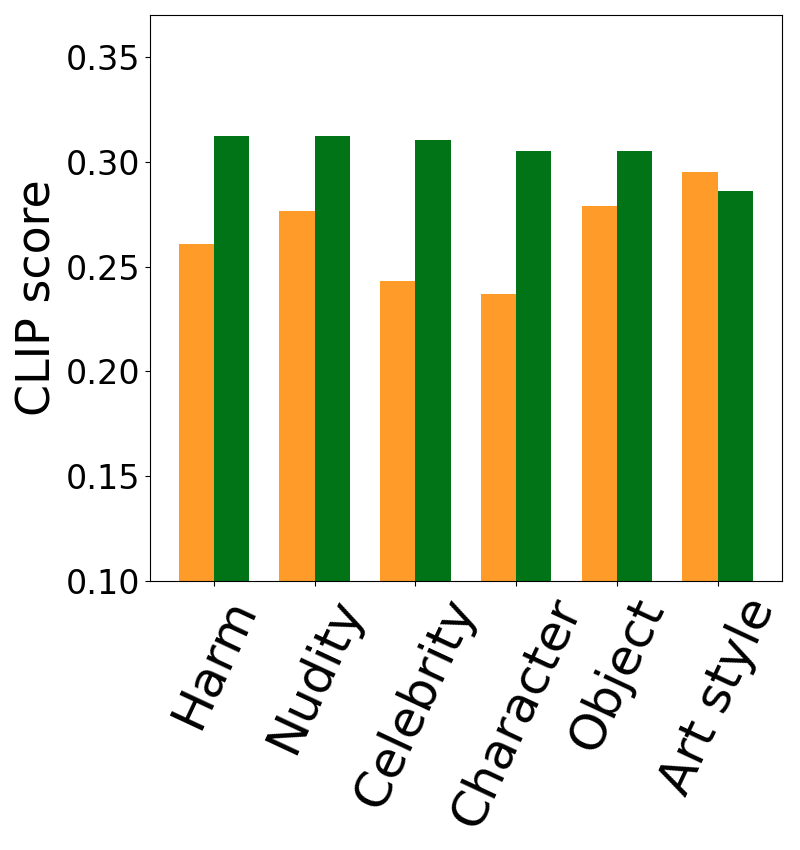}
        \vspace{-0.21in}
        \caption{SLD}
        \label{fig:clip_score_9_SLD}
    \end{subfigure}\hfill
    \begin{subfigure}{0.19\textwidth}
        \centering
        \includegraphics[width=\textwidth]{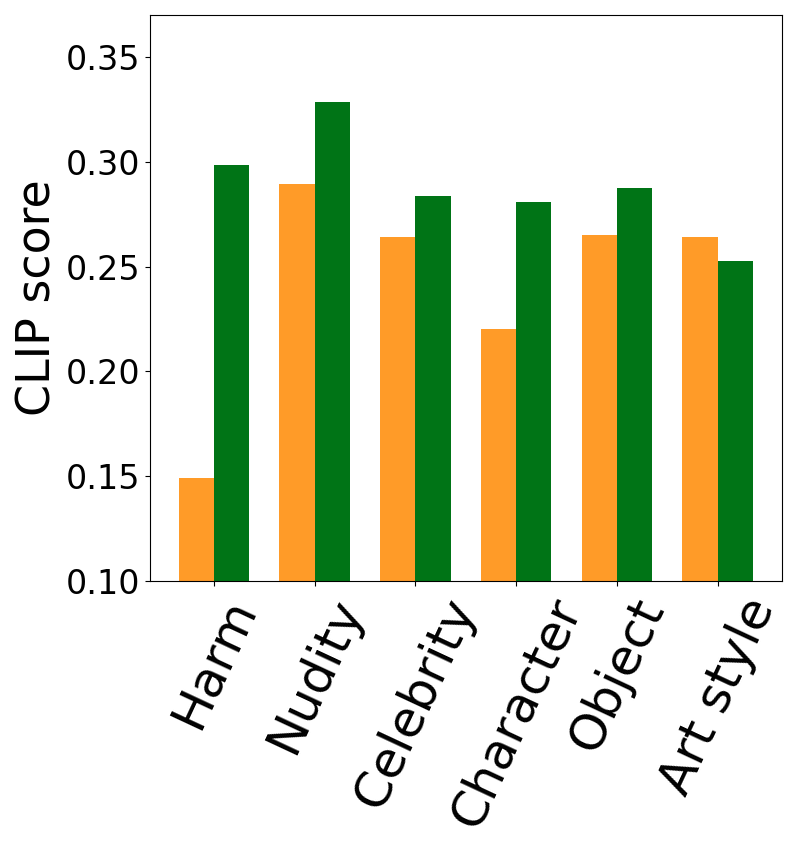}
        \vspace{-0.21in}
        \caption{SEGA}
        \label{fig:clip_score_X_SEGA}
    \end{subfigure}
    \caption{In-prompt retainability vs. out-prompt retainability}
    \label{fig:retainability}
    \vspace{-0.1in}
\end{figure*}
We evaluate the generation ability on benign concepts using both in-prompt and out-prompt CLIP scores. In Fig.~\ref{fig:retainability}, we use DVD to test the in-prompt CLIP score and a subset with benign prompts from LAION~\cite{schuhmann2022laion} to test out-prompt CLIP score. The observations as follows.

\textbf{Observation 1: In-prompt retainability performs worse than out-prompt retainability.} As shown in Fig.~\ref{fig:retainability}, the in-prompt CLIP score is lower than the out-prompt CLIP score across all the methods and concept categories. This indicates that the ability to generate benign parts in prompts containing malicious concepts is more negatively affected than the ability to generate totally benign prompts. As mentioned in Sec.~\ref{sec:in_prompt}, the in-prompt retainability is also important for concept removal methods. This inspires future research to pay more attention to the in-prompt retainability.
\textbf{\textit{Remark.}} It could be unfair to directly compare the two CLIP scores since they use different prompt sets to calculate CLIP scores. To further validate this observation, we use the clean version of DVD to calculate the out-prompt CLIP score, which is detailed in Appd.~\ref{appd:out_prompt}. By comparing it with the in-prompt CLIP score in Appd.~\ref{appd:out_prompt}, we have observed a consistent phenomenon with this subsection.


\textbf{Observation 2: Auxiliary semantic information in MACE (Fig.~\ref{fig:clip_score_7_MACE}) may hurt the generation on benign concepts.} 
MACE has the most significant decrease in the generation ability of benign concepts for both in-prompt and out-prompt CLIP scores. We conjecture that this is because it introduces auxiliary semantic information to help locate the unwanted concepts in the image. In other methods, the unwanted concepts are usually located by the semantic understanding ability of the T2I model itself. In contrast, MACE incorporates the segmentation results of Grounded-SAM~\cite{liu2023grounding, kirillov2023segment} to locate the image area of unwanted concepts. It focuses on optimizing this area and overlooks other areas of the image. This may lead to worse retainability in other areas. Another piece of evidence is that when we do not include segmentation in removing the art style concepts by MACE (since art style is a global feature), it has good in-prompt and out-prompt CLIP scores. 


\section{Conclusion and Ethical Statement}
\label{sec:conclusion}
In this work, we address the lack of a systematic benchmark for concept removal methods by introducing a comprehensive and effective dataset and a new evaluation metric. Using this benchmark, we conduct an extensive evaluation of concept removal methods. Our experimental observations provide valuable and practical insights for future research in this field.
While our benchmark may have a limitation in the inconvenient detection metric for art styles (using the CLIP score for art styles, instead of the detection rate from a classifier like other categories), it still offers a thorough and practical evaluation across the various method settings.

\textbf{Ethical statement.} This work provides a comprehensive evaluation framework for concept removal, which is targeted to mitigate the potential malicious use of text-to-image diffusion models. Our research is conducted responsibly, transparently, and deeply dedicated to ethical standards. Despite involving the generation of sensitive content such as nudity and violence, it is strictly for research purposes and does not intend to produce or promote inappropriate material. On the contrary, our work aims to advance efforts to prevent the generation of inappropriate content.

\section*{Acknowledgement}
Jie Ren, Kangrui Chen, Yingqian Cui, Shenglai Zeng, and Jiliang Tang are supported by the National Science Foundation (NSF) under grant numbers CNS2321416, IIS2212032, IIS2212144, IOS2107215, DUE2234015, CNS2246050, 
\newline DRL2405483 and IOS2035472, the Army Research Office (ARO) under grant number W911NF-21-1-0198, Amazon Faculty Award, JP Morgan Faculty Award, Meta, Microsoft and SNAP.
{
    \small
    \bibliographystyle{ieeenat_fullname}
    \bibliography{main}
}

\clearpage
\setcounter{page}{1}
\maketitlesupplementary

\appendix

\section{Documentation of the Proposed Datasets}
\label{appd:dataset}

\subsection{Six-CD}
\label{appd:six_cd}

In Six-CD, we provide six categories of concepts to test concept removals. For the two general categories, we provide 991 effective prompts for harm concept and 1539 effective prompts for nudity concept. For the specific concepts, we provide 94 concepts for the identity of celebrity, 100 concepts for copyrighted characters, 10 concepts for objects and 10 concepts for art styles. All the prompts, concepts and templates for specific concepts are attached in the supplementary materials.


\subsection{Dual-Version Dataset}
\label{appd:dual_prompt}

In Dual-Version Dataset, for each category, we provide a malicious version and a clean version. The two versions are documented in separate files. For specific concepts, the two versions are constructed by the templates generated by ChatGPT and the concepts in Six-CD.
The templates for specific concepts are provided in an extra file. 

\section{Baseline Settings}
\label{appd:baseline}

We use the official code provided by the respective papers for all baselines. In some categories of SPM and MACE, we utilize the officially released checkpoints. For categories with provided hyper-parameters, we use those directly. For other categories without specified hyper-parameters, we fine-tune the learning rate, training steps, and other specific parameters of the method.

\begin{figure}[h]
    \centering
    \includegraphics[width=\linewidth]{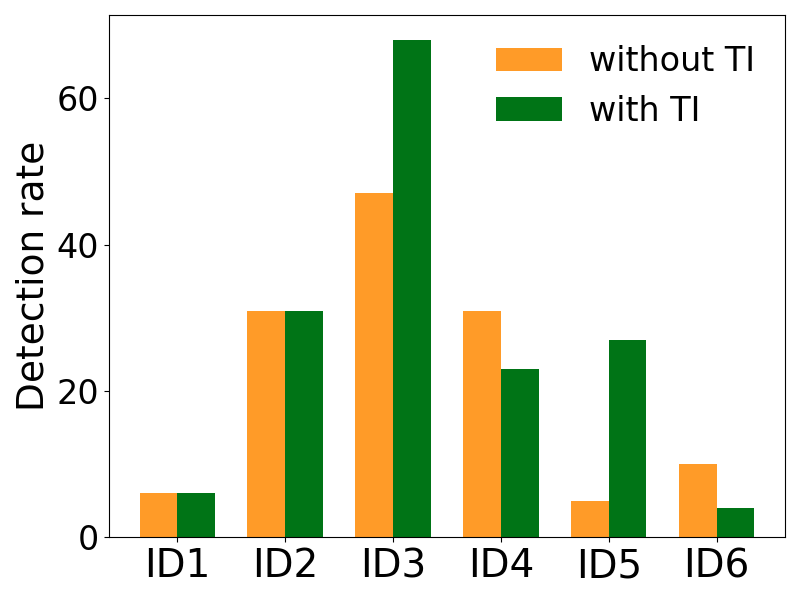}
    \caption{FMN with and without TI. We test six identities of celebrities in FMN (FMN is majorly used to remove celebrities in the original paper~\cite{zhang2023forget}). The results show that TI is a random factor. For some identities, such as ID4 and ID6, it has the positive influence on the removal ability, while for some identities such as ID3 and ID5, it has negative influence on the removal ability. }
    \label{fig:fmn}
\end{figure}
For ESD, we use the variant ESD-x for art styles and the variant ESD-u for others, which is consistent with original paper.
For FMN, we test the removal ability with and without Textual Inversion (TI) and find they have similar performance, which is shown in Fig.~\ref{fig:fmn}. In our benchmark, we use FMN without TI for all the experiments. Also, FMN is not suitable for multiple concepts since it requires massive collection of images for each concept. Thus, we exclude it for multiple concepts.

\section{License of Assets}
\label{appd:license}

\begin{table*}[t]
    \centering
   \caption{License information of assets}
    \resizebox{\linewidth}{!}{ 
    \begin{tabular}{lll}
        \toprule
        Asset & License & Link \\
        \midrule
        I2P & MIT license & https://huggingface.co/datasets/AIML-TUDA/i2p \\
        MMA & cc-by-nc-nd-3.0 & https://huggingface.co/datasets/YijunYang280/MMA-Diffusion-NSFW-adv-prompts-benchmark \\
        SD-uncensored & MIT license & https://huggingface.co/datasets/jtatman/stable-diffusion-prompts-stats-full-uncensored \\
        UD & Not found & https://github.com/YitingQu/unsafe-diffusion \\
        CPDM & Not found & https://arxiv.org/abs/2403.12052v1 \\
        GCD & MPL-2.0 license & https://github.com/Giphy/celeb-detection-oss \\
        \midrule
        NEG & creativeml-openrail-m & https://huggingface.co/CompVis/stable-diffusion-v1-4 \\
        ESD & MIT license & https://github.com/rohitgandikota/erasing \\
        SPM & Apache-2.0 license & https://github.com/Con6924/SPM \\
        SDD & MIT license & https://github.com/nannullna/safe-diffusion \\
        FMN & MIT license & https://github.com/SHI-Labs/Forget-Me-Not \\
        UCE & MIT license & https://github.com/rohitgandikota/unified-concept-editing \\
        MACE & MIT license & https://github.com/Shilin-LU/MACE \\
        EMCID & MIT license & https://github.com/SilentView/EMCID/tree/master \\
        SLD & MIT license & https://github.com/ml-research/safe-latent-diffusion \\
        SEGA & MIT license & https://github.com/ml-research/semantic-image-editing \\
        NudeNet & AGPL-3.0, AGPL-3.0 licenses found & https://github.com/notAI-tech/NudeNet \\
        Q16 & Not found & https://github.com/ml-research/Q16 \\
        \bottomrule
    \end{tabular}
    }
    \label{tab:license}
\end{table*}

In Table~\ref{tab:license}, we present the license information of all the assets including the data resources collected for the concepts and the code for all the concept removal methods and detection methods we use in this paper.

\section{Additional Experiments}
\label{appd:exp}

\subsection{Human Evaluation of Proposed New Metric}
\label{appd:human}

We provide additional human evaluations Table~\ref{tab:human} to demonstrate the validity of the proposed metric (in-prompt CLIP score). 

Settings: 
We evaluate the score based on 50 pairs of images. Each pair of images are generated using the same malicious prompt and using two different concept removal methods. In addition to the malicious prompt, we use the corresponding clean version of the malicious prompt in Dual-Version Dataset. Given the pair of images and the clean prompt, we ask humans to rank which image better aligns with the clean prompt. After collecting the ranking results from 15 humans, we compare the ranking with the order given by the in-prompt CLIP scores of the two images. If the human preference is the same as the CLIP score, we label the sample as correct. With 15 humans, there are a total of 750 comparisons done in our evaluation.

Results: We report the accuracy in the following table. Note that if the two CLIP scores are close to each other, the human preference may be inaccurate. Therefore, we report the results when the difference is larger than a threshold $T$. From the table below, we can see that when $T=0$, the accuracy is 86.86\%, which means most of the in-prompt CLIP score is consistent with human preference. When $T=0.03$, the correctness is even larger than 95\%. These observations validate the effectiveness of our proposed metric.

\begin{table}[!t]
    \centering
    \caption{Human Evaluation of in-prompt CLIP score}
    \begin{tabular}{llll}
    \toprule
        $T$ & 0 & 0.025 & 0.03 \\ \midrule
        Acc. & 86.86\% & 91.24\% & 95.24\% \\ \bottomrule
    \end{tabular}
    \label{tab:human}
\end{table}

\subsection{Table of Removal Ability with Error Bar}
\label{appd:error_bar}
Besides Fig.~\ref{fig:main}, we also report the results of removal ability and the error bar in Table~\ref{tab:error_bar}. 
We calculate the standard variance using the estimation of bootstrap\footnote{https://wires.onlinelibrary.wiley.com/doi/full/10.1002/wics.182, https://ieeexplore.ieee.org/abstract/document/4767957, https://yuleii.github.io/2021/01/22/bootstrap.html}.

\begin{table*}[t]
    \centering
   \caption{Removal ability}
    \resizebox{1\textwidth}{!}{ 
    \begin{tabular}{lcccccc}
        \toprule
        & Harm & Nudity & Celebrity & Character & Object & Art style \\
        \midrule
        V1-4 & 0.7683 {\small $\pm$ 0.0174} &
            0.8096 {\small $\pm$ 0.0129} &
            0.9407 {\small $\pm$ 0.0012} &
            0.9704 {\small $\pm$ 0.0087} &
            0.9335 {\small $\pm$ 0.0128} &
            0.3144 {\small $\pm$ 0.0012} \\
        
        NEG & 0.4546 {\small $\pm$ 0.0202} &
            0.2075 {\small $\pm$ 0.0134} &
            0.2415 {\small $\pm$ 0.0222} &
            0.1758 {\small $\pm$ 0.0197} &
            0.4497 {\small $\pm$ 0.0259} &
            0.2761 {\small $\pm$ 0.0018} \\
        
        ESD & 0.5072 {\small $\pm$ 0.0204} &
            0.1195 {\small $\pm$ 0.0107} &
            0.0224 {\small $\pm$ 0.0076} &
            0.0064 {\small $\pm$ 0.0040} &
            0.0815 {\small $\pm$ 0.0142} &
            0.2237 {\small $\pm$ 0.0028} \\
        
        SPM & 0.7689 {\small $\pm$ 0.0172} &
            0.8032 {\small $\pm$ 0.0132} &
            0.0360 {\small $\pm$ 0.0096} &
            0.1162 {\small $\pm$ 0.0164} &
            0.5031 {\small $\pm$ 0.0259} &
            0.3064 {\small $\pm$ 0.0014} \\
        
        SDD & 0.2023 {\small $\pm$ 0.0164} &
            0.0376 {\small $\pm$ 0.0062} &
            0.2546 {\small $\pm$ 0.0222} &
            0.0407 {\small $\pm$ 0.0101} &
            0.1741 {\small $\pm$ 0.0197} &
            0.2791 {\small $\pm$ 0.0016} \\
        
        FMN & 0.7238 {\small $\pm$ 0.0181} &
            0.7991 {\small $\pm$ 0.0131} &
            0.3055 {\small $\pm$ 0.0238} &
            0.1391 {\small $\pm$ 0.0179} &
            0.7033 {\small $\pm$ 0.0236} &
            0.2826 {\small $\pm$ 0.0016} \\
        
        UCE & 0.5355 {\small $\pm$ 0.0204} &
            0.1051 {\small $\pm$ 0.0099} &
            0.0016 {\small $\pm$ 0.0020} &
            0.0199 {\small $\pm$ 0.0072} &
            0.0982 {\small $\pm$ 0.0152} &
            0.2488 {\small $\pm$ 0.0020} \\
        
        MACE & 0.2708 {\small $\pm$ 0.0185} &
            0.0370 {\small $\pm$ 0.0062} &
            0.0247 {\small $\pm$ 0.0079} &
            0.0000 {\small $\pm$ 0.0000} &
            0.0720 {\small $\pm$ 0.0133} &
            0.2670 {\small $\pm$ 0.0020} \\
        
        EMCID & 0.7685 {\small $\pm$ 0.0174} &
            0.8063 {\small $\pm$ 0.0130} &
            0.3398 {\small $\pm$ 0.0242} &
            0.2943 {\small $\pm$ 0.0235} &
            0.6200 {\small $\pm$ 0.0247} &
            0.3141 {\small $\pm$ 0.0012} \\
        
        SLD & 0.3142 {\small $\pm$ 0.0189} &
            0.4166 {\small $\pm$ 0.0162} &
            0.0040 {\small $\pm$ 0.0032} &
            0.0815 {\small $\pm$ 0.0140} &
            0.0856 {\small $\pm$ 0.0143} &
            0.2279 {\small $\pm$ 0.0021} \\
        
        SEGA & 0.1689 {\small $\pm$ 0.0154} &
            0.5361 {\small $\pm$ 0.0163} &
            0.8728 {\small $\pm$ 0.0173} &
            0.9288 {\small $\pm$ 0.0131} &
            0.8894 {\small $\pm$ 0.0161} &
            0.3107 {\small $\pm$ 0.0013} \\
        
        \bottomrule
    \end{tabular}
    }
    \label{tab:error_bar}
\end{table*}

\subsection{Out-prompt CLIP Score by Clean Version of DVD}
\label{appd:out_prompt}

\begin{figure*}[t]
    \centering
    \begin{subfigure}{0.98\textwidth}
        \centering
        \includegraphics[width=0.4\textwidth]{pics/clip_score_legend.png}
        \label{fig:appd_clip_score_legend}
    \end{subfigure}\hfill
    \begin{subfigure}{0.19\textwidth}
        \centering
        \includegraphics[width=\textwidth]{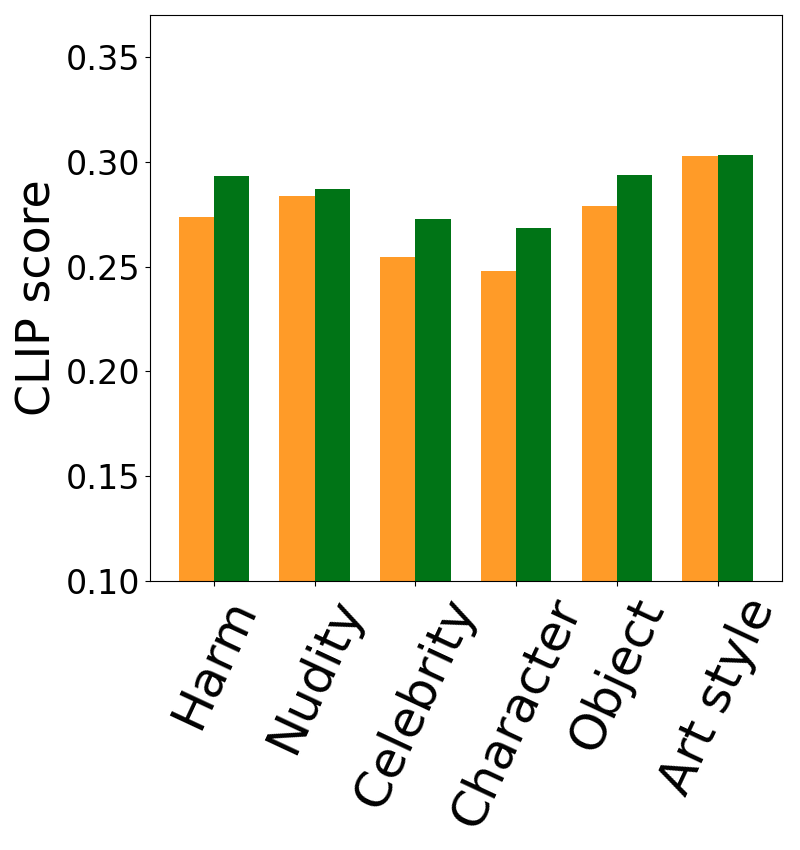}
        \vspace{-0.21in}
        \caption{NEG}
        \label{fig:appd_clip_score_1_NEG}
    \end{subfigure}\hfill
    \begin{subfigure}{0.19\textwidth}
        \centering
        \includegraphics[width=\textwidth]{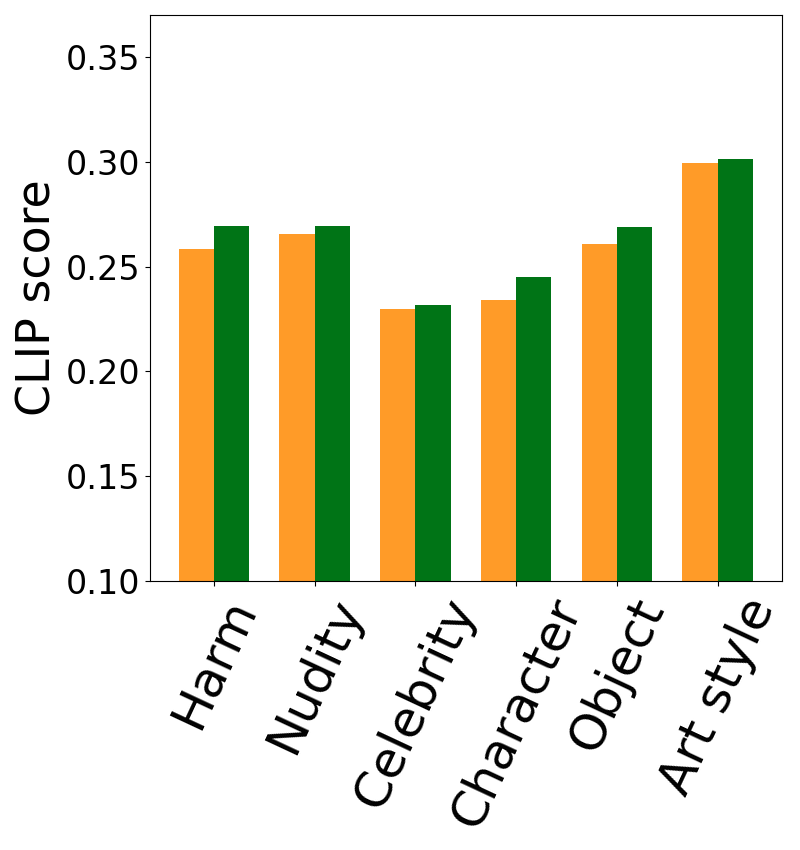}
        \vspace{-0.21in}
        \caption{ESD}
        \label{fig:appd_clip_score_2_ESD}
    \end{subfigure}\hfill
    \begin{subfigure}{0.19\textwidth}
        \centering
        \includegraphics[width=\textwidth]{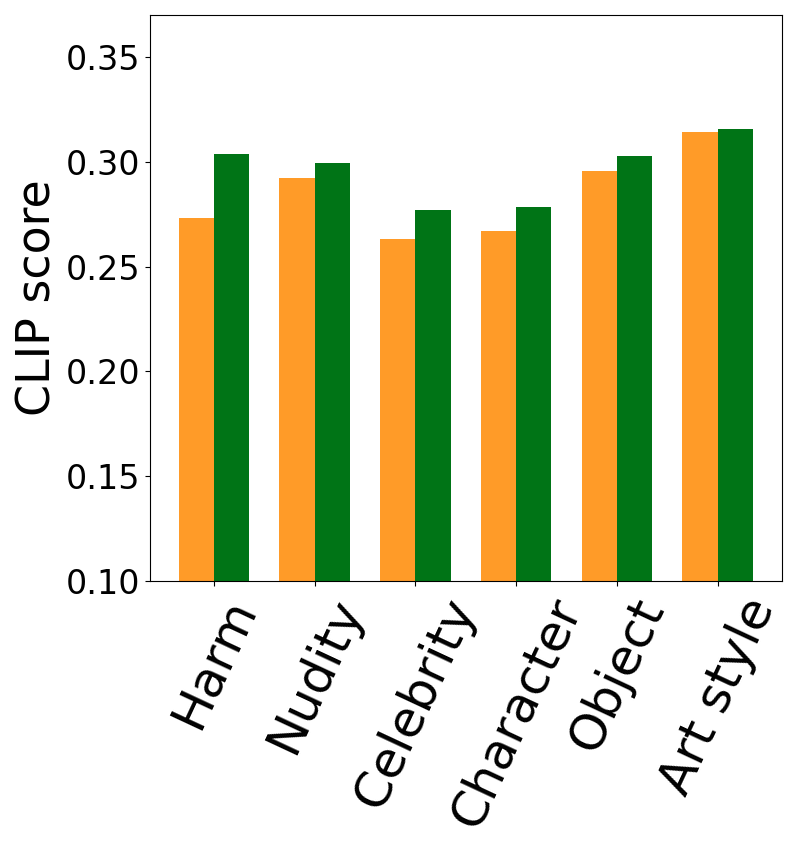}
        \vspace{-0.21in}
        \caption{SPM}
        \label{fig:appd_clip_score_3_SPM}
    \end{subfigure}\hfill
    \begin{subfigure}{0.19\textwidth}
        \centering
        \includegraphics[width=\textwidth]{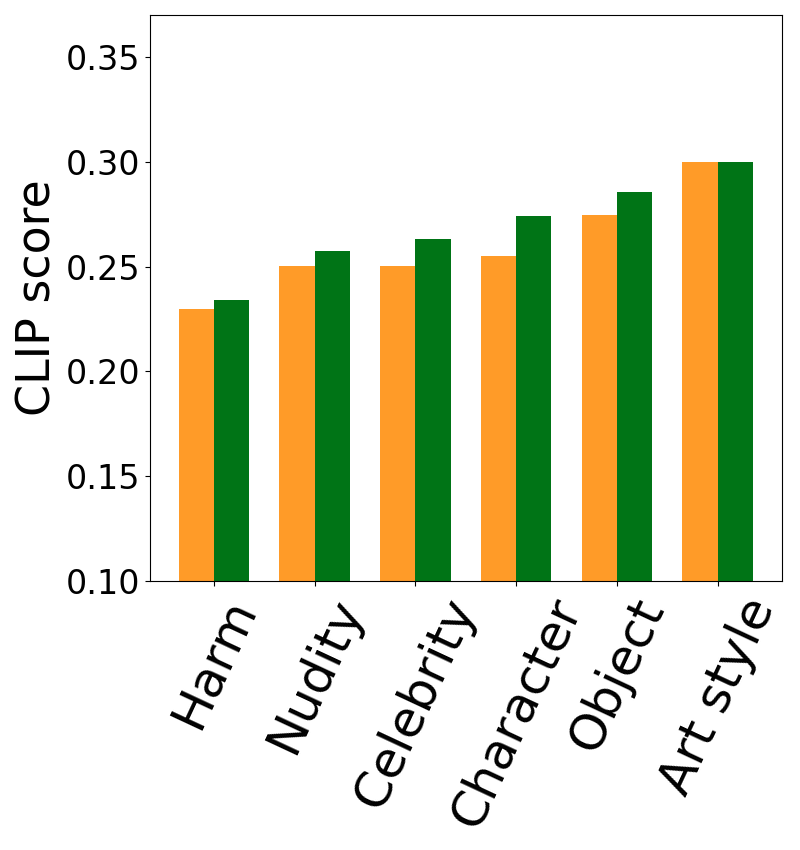}
        \vspace{-0.21in}
        \caption{SDD}
        \label{fig:appd_clip_score_4_SDD}
    \end{subfigure}\hfill
    \begin{subfigure}{0.19\textwidth}
        \centering
        \includegraphics[width=\textwidth]{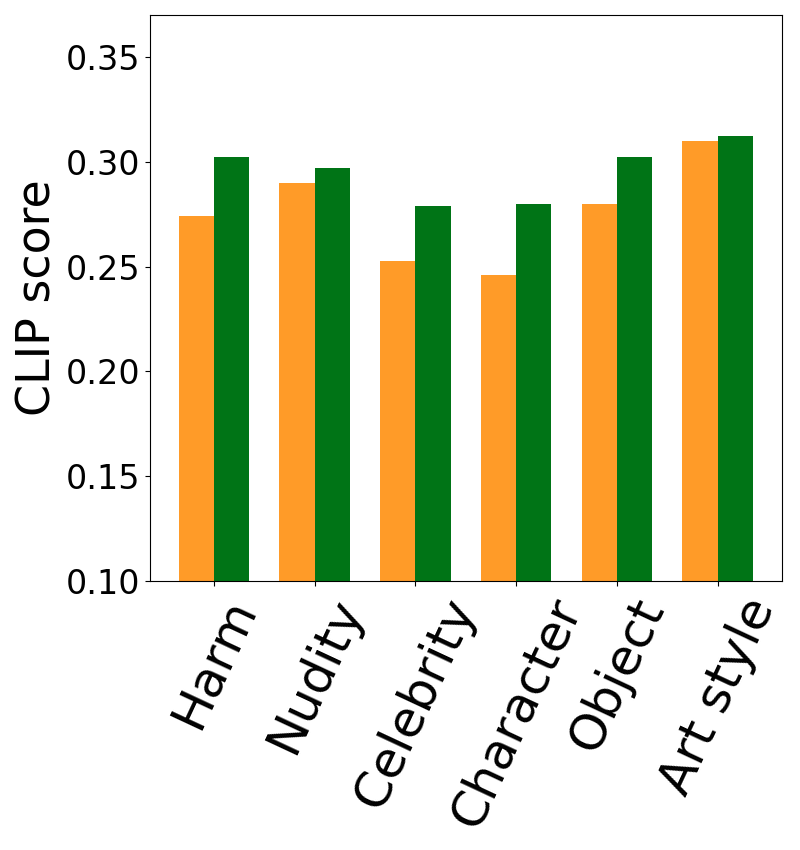}
        \vspace{-0.21in}
        \caption{FMN}
        \label{fig:appd_clip_score_5_FMN}
    \end{subfigure}\hfill
    \begin{subfigure}{0.19\textwidth}
        \centering
        \includegraphics[width=\textwidth]{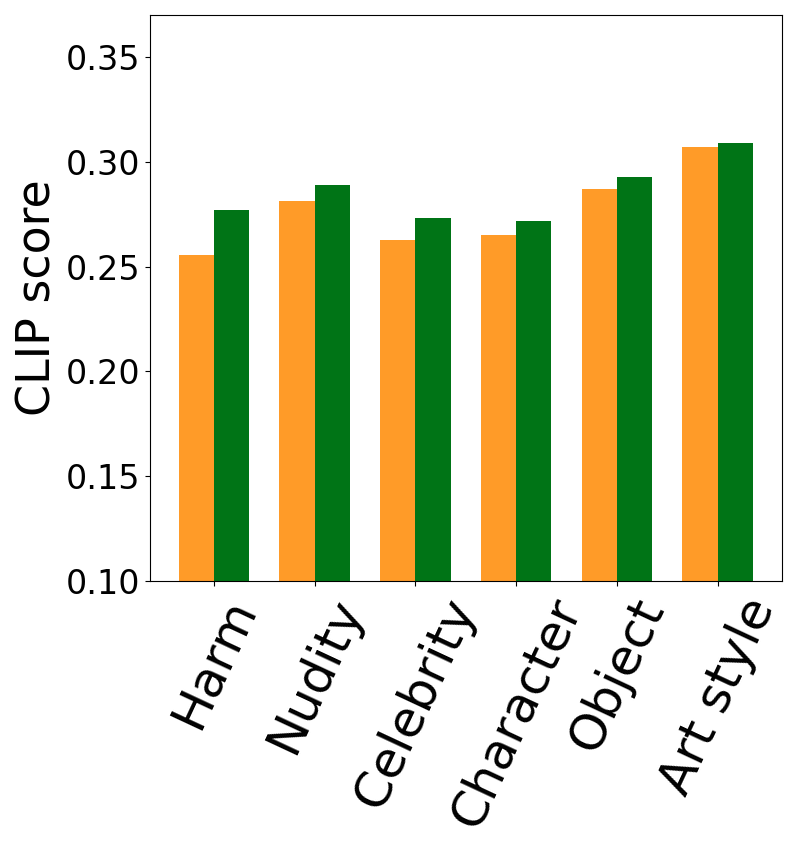}
        \vspace{-0.21in}
        \caption{UCE}
        \label{fig:appd_clip_score_6_UCE}
    \end{subfigure}\hfill
    \begin{subfigure}{0.19\textwidth}
        \centering
        \includegraphics[width=\textwidth]{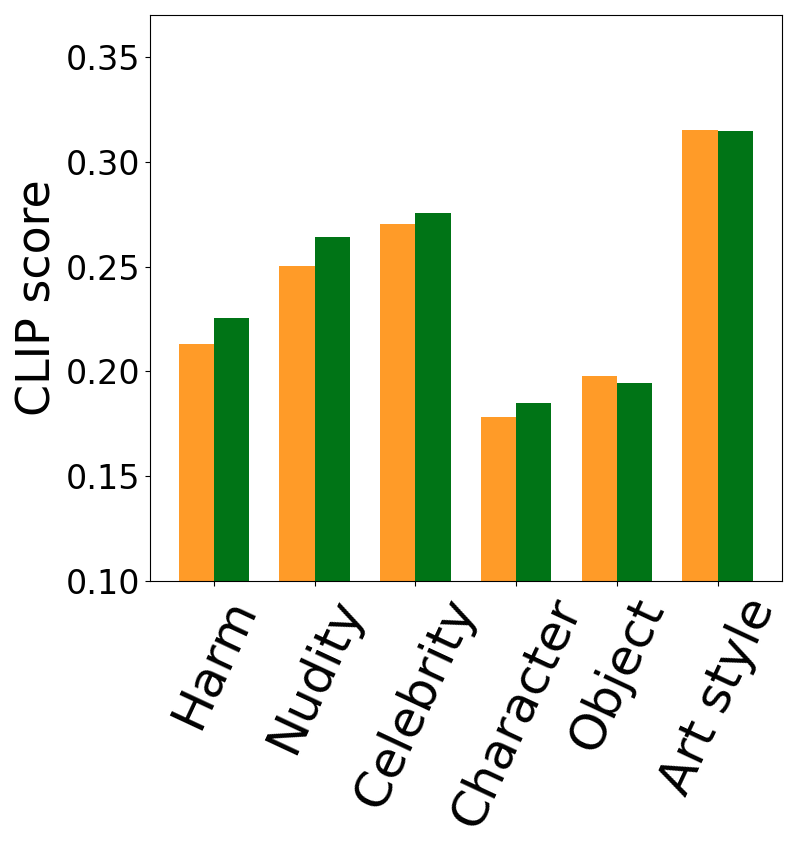}
        \vspace{-0.21in}
        \caption{MACE}
        \label{fig:appd_clip_score_7_MACE}
    \end{subfigure}\hfill
    \begin{subfigure}{0.19\textwidth}
        \centering
        \includegraphics[width=\textwidth]{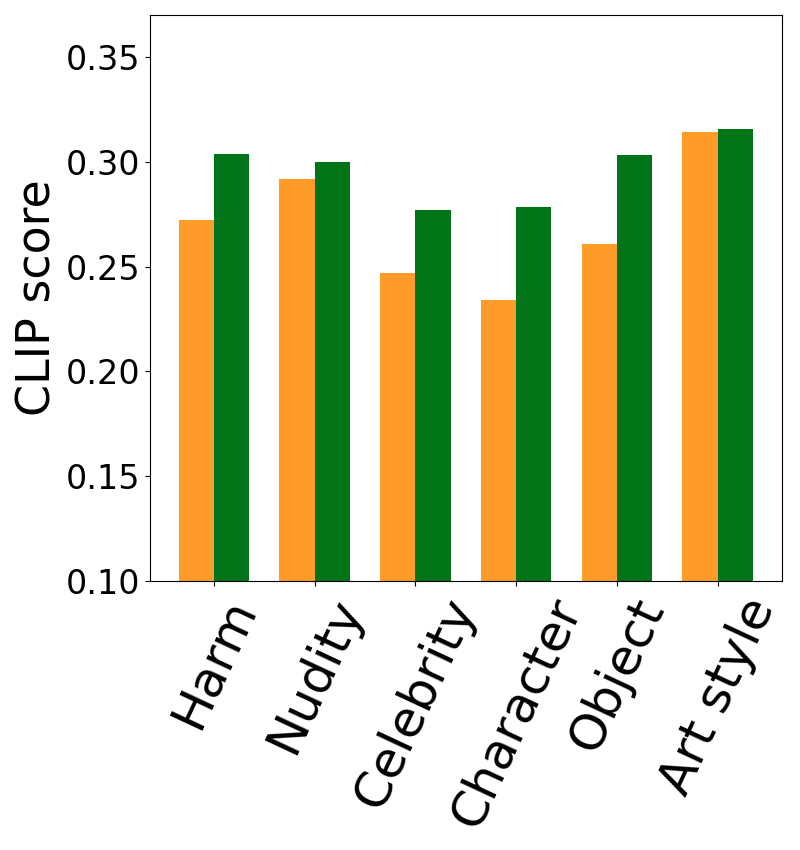}
        \vspace{-0.21in}
        \caption{EMCID}
        \label{fig:appd_clip_score_8_EMCID}
    \end{subfigure}\hfill
    \begin{subfigure}{0.19\textwidth}
        \centering
        \includegraphics[width=\textwidth]{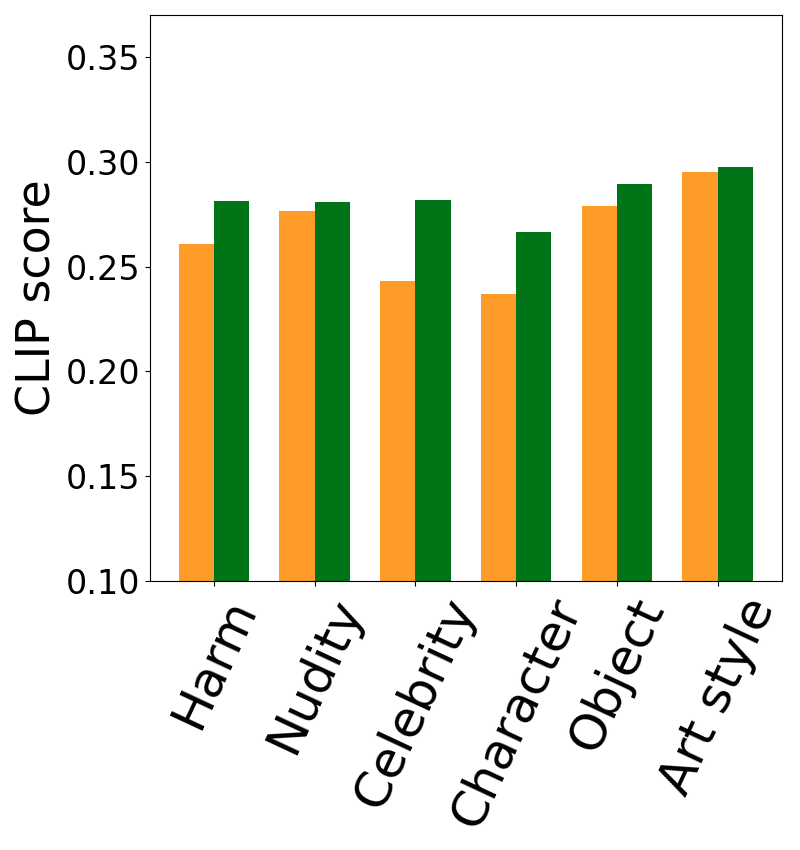}
        \vspace{-0.21in}
        \caption{SLD}
        \label{fig:appd_clip_score_9_SLD}
    \end{subfigure}\hfill
    \begin{subfigure}{0.19\textwidth}
        \centering
        \includegraphics[width=\textwidth]{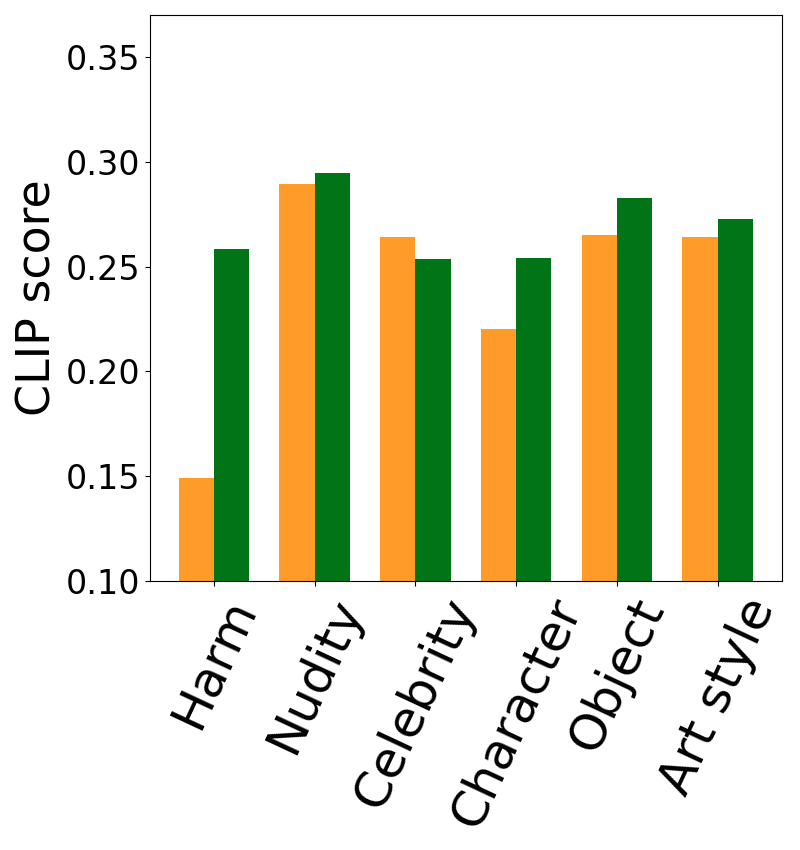}
        \vspace{-0.21in}
        \caption{SEGA}
        \label{fig:appd_clip_score_X_SEGA}
    \end{subfigure}
    \caption{In-prompt retainability vs. out-prompt retainability by clean version of DVD}
    \label{fig:retainability_appd}
\end{figure*}
We use the clean version prompt of DVD to calculate the out-prompt CLIP score in Fig.~\ref{fig:retainability_appd} and get the consistent conclusion with Sec.~\ref{exp:in_prompt_retainability_53}. As we can see, the out-prompt CLIP score is still higher than in-prompt CLIP score in almost all the concepts and removal methods. This means concept removals will have more severe impact on the in-prompt retainability than the retainability on the totally benign prompts. Thus, in the design of concept removals, in-prompt retainability should be considered carefully.

\subsection{Time Cost}
\label{appd:time}

\begin{table}[t]
    \centering
   \caption{Time cost of training and inference}
    \begin{tabular}{lcccc}
        \toprule
        & \multicolumn{2}{c}{Training} & \multicolumn{2}{c}{Inference} \\
        & Single & Multiple  & Single & Multiple \\
        \midrule
        NEG & N/A & N/A & 7.05s & 7.09s \\
        ESD & 69.18m & 67.38m & 6.08s & 6.09s \\
        SPM & 152.64m & 254.09h & 9.11s & 9.39s \\
        SDD & 96.76m & 97.38m & 7.74s & 7.81s \\
        UCE & 0.15m & 0.80m & 7.68s & 7.73s \\
        MACE & 1.80m & 64.00m & 7.14s & 7.19s \\
        EMCID & 1.16m & 112.53m & 7.81s & 7.81s \\
        SLD & N/A & N/A & 10.33s & 10.38s \\
        SEGA & N/A & N/A & 10.46s & OOM \\
        \bottomrule
    \end{tabular}
    \label{tab:time_cost}
\end{table}

We show the training and inference time of all the methods in Table~\ref{tab:time_cost}. 

For training time cost, we train each method for single concept removal and for multiple removal. In multiple concept removal, we remove 100 concepts. The experiments are conducted on A5000 (except MACE of multiple removal that is trained on A6000 due to OOM). As we can see, some methods, such as ESD, UCE and SDD, have similar training time in single and multiple. It means the training time will not increase as the number of concepts increase. But other methods have significantly increased time in multiple concepts compared with single concepts.

For inference time cost, we test the time cost to generate one image on A5000. We can see that, most of methods have similar inference time cost at around 7 seconds. However, SPM, SLD and SEGA may have increased inference time. SEGA causes OOM on A5000 when removing 100 concepts. We test its time to generate one image when removing 50 concepts, which is 170 seconds. Thus, when the number of removed concept is increasing, SEGA increases the requirement of both GPU memory and inference time cost.

\subsection{Fine-grained Retainability for Similar Concepts}
\label{appd:fine_grained}

\begin{figure*}[t]
    \centering
    \begin{subfigure}{0.32\textwidth}
        \centering
        \includegraphics[width=\textwidth]{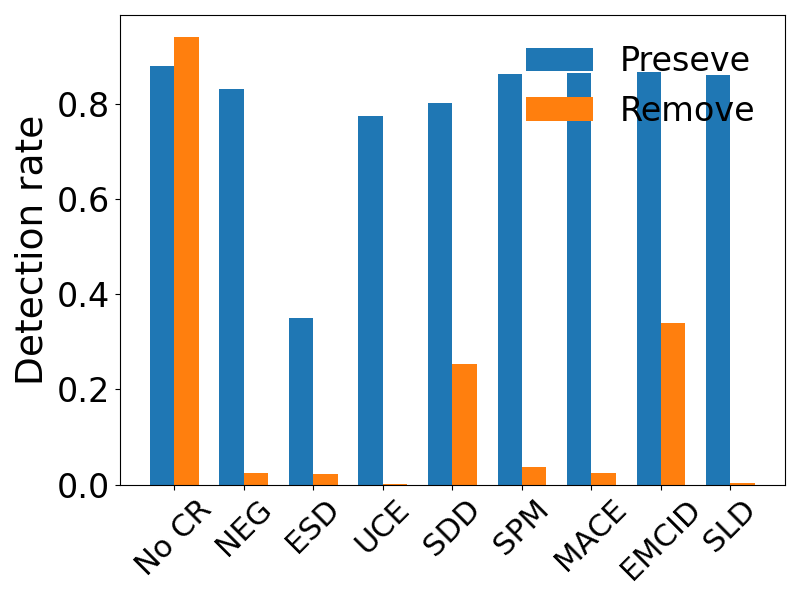}
        \caption{Removing single concept}
        \label{fig:preserve_1}
    \end{subfigure}
    \hfill
    \begin{subfigure}{0.32\textwidth}
        \centering
        \includegraphics[width=\textwidth]{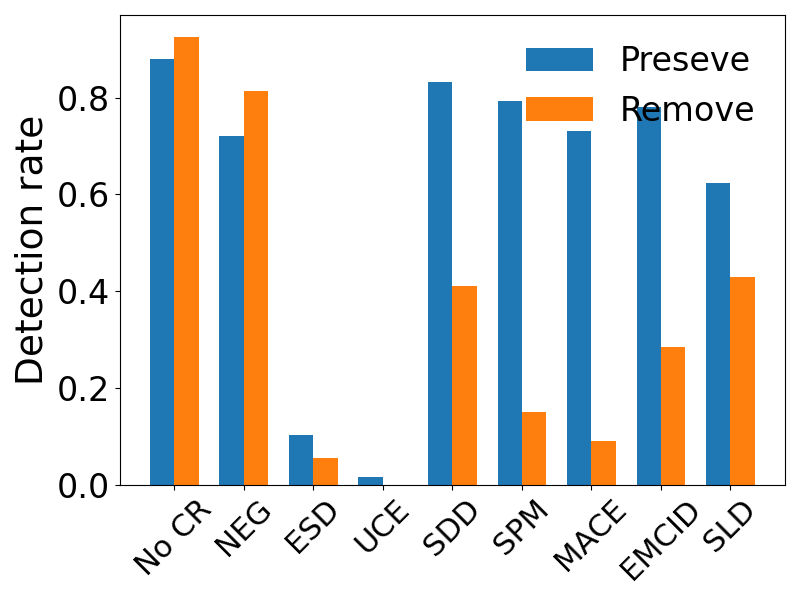}
        \caption{Removing 10 concept}
        \label{fig:preserve_10}
    \end{subfigure}
    \hfill
    \begin{subfigure}{0.32\textwidth}
        \centering
        \includegraphics[width=\textwidth]{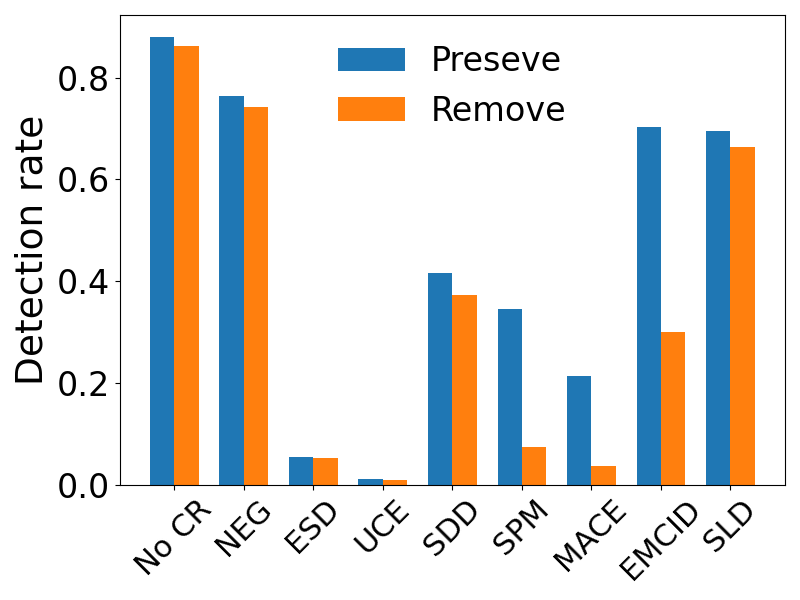}
        \caption{Removing 50 concept}
        \label{fig:preserve_100}
    \end{subfigure}
    \caption{Influence on similar concepts}
    \label{fig:similar}
\end{figure*}
When removing concepts from diffusion models, similar benign concepts are more likely to be influenced. For example, when removing certain celebrities, other celebrities not included in the removal set may also be affected. Therefore, we test retainability on similar concepts. In Fig.~\ref{fig:similar}, we remove 1/10/50 celebrity concepts in Six-CD and preserve the generation ability on other 44 celebrity concepts. When removing a single concept, the generation ability on the preserved concepts remains strong, except for ESD. However, when the number of removed concepts increases to 10, the generation abilities of ESD and UCE on preserved concepts significantly decrease. When the number of removed concepts reaches 50, only EMCID, NEG and SLD perform well on the preserved concepts, but EMCID's ability to remove the concepts is also worse than the others, while ENG and SLD have almost no ability in removing multiple concepts. This means the ability to preserve similar concepts for all the methods still requires improvement.

\subsection{FID}
\label{appd:fid}

\begin{figure*}[t]
    \centering
    \begin{subfigure}{0.40\textwidth}
        \centering
        \includegraphics[width=\textwidth]{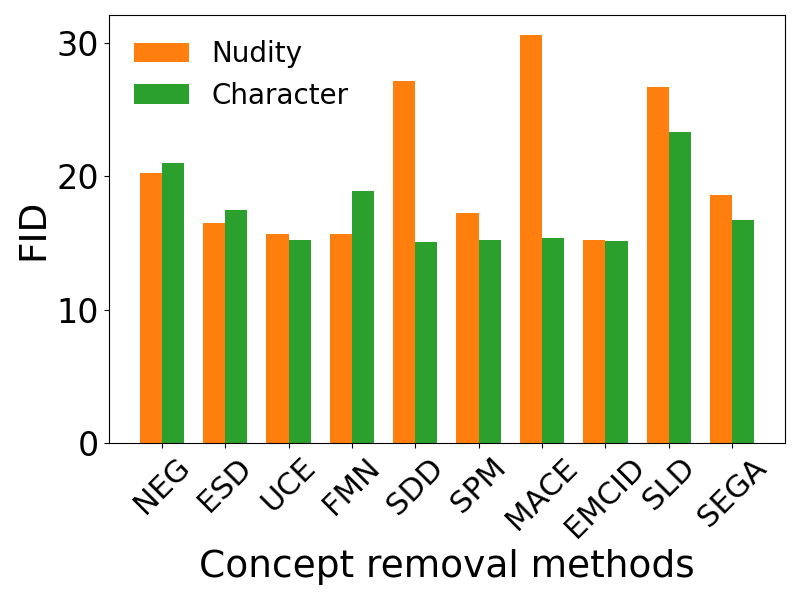}
        \caption{Removing single concept}
        \label{fig:fid_single}
    \end{subfigure}
    \hspace{0.1in}
    \begin{subfigure}{0.40\textwidth}
        \centering
        \includegraphics[width=\textwidth]{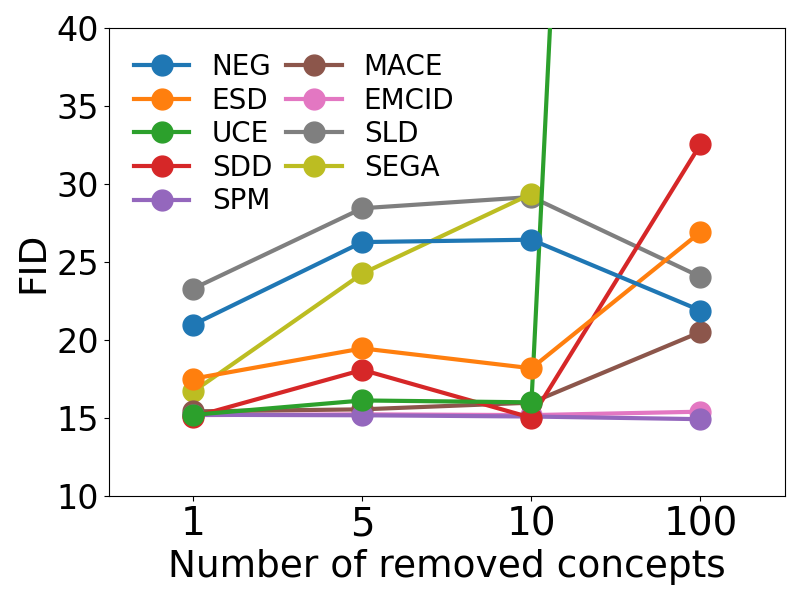}
        \caption{Removing multiple concept}
        \label{fig:fid_multiple}
    \end{subfigure}
    \caption{FID}
    \label{fig:fid}
\end{figure*}
We test FID on nudity concept and character concept, which are two representative categories from general and specific concepts in Fig.~\ref{fig:fid_single}. We also plot the trend of FID as the number of concept increases in Fig.~\ref{fig:fid_multiple}. In single concept removal, most methods show similar performance. However, for inference-time methods, the FIDs for both nudity and character concepts are higher than those of other methods, indicating that inference-time mitigation is too aggressive and negatively impacts generation quality. Additionally, for the nudity concept, MACE and SDD exhibit significantly worse FID scores compared to others. In multiple concept removal, only EMCID and SPM maintain the generation quality when removing 100 concepts. In contrast, UCE performs poorly, with a significantly increased FID.

\subsection{Experiments on other models}
\label{appd:other_model}

Besides SD v1.4, we provide additional experiments on DPO-Diffusion and SD v1.5 in this section. The observations in this section are consistent with the findings in our paper:

\begin{itemize}
    \item General concepts are harder to remove. In the following Table~\ref{tab:appd_gen_vs_spec}, we choose Nudity and Copyright to represent general and specific concepts, respectively. Similar to our findings in Section 5.1, after concept removals, the generated general concepts are higher than specific concepts in different methods.
    \item Inference-time methods fail in removing multiple concepts. In the following Table~\ref{tab:appd_multi}, the two inference-time methods, NEG and SLD, have poor ability in removing unwanted concepts when the number of concepts increases to 100. To explain this, as mentioned in Section 5.2, these methods have to encode the string containing all the concepts in the embeddings of one single prompt. The long string will exceed the capacity of the text encoder of T2I diffusion models and lead to failed removal.
    \item Closed-form solutions by modifying linear components perform well in removing multiple concepts. As shown in the following Table~\ref{tab:appd_multi}, UCE performs well in removing 100 concepts: only 0.75\% of images are detected with unwanted concepts, while NEG is 81.05\%. We conjecture that combining and changing multiple concepts in the linear components is easier than in other non-linear parts.
    \item In-prompt retainability performs worse than out-prompt retainability. In the following Table~\ref{tab:appd_clip_score}, the in-prompt CLIP score is lower than the out-prompt CLIP score across different methods and concept categories. This result is also consistent with our observation in Section 5.3.
\end{itemize}

\begin{table}[!t]
    \caption{Detection rate of nudity and copyright concepts after concept removals (model: DPO-Diffusion).}
    \centering
    \begin{tabular}{llllll}
    \toprule
        & no CR & NEG & EMCID & SDD & UCE \\ \midrule
        Nudity & 0.8960 & 0.3258 & 0.8937 & 0.0828 & 0.1482 \\ 
        Copy. & 0.9736 & 0.2560 & 0.3088 & 0.0048 & 0.0152 \\ \bottomrule
    \end{tabular}
    \label{tab:appd_gen_vs_spec}
\end{table}

\begin{table}[!t]
\caption{Detection rate after removing multiple concepts (model: SD v1.5).}
    \centering
    \begin{tabular}{llll}
    \toprule
        Copyright number & NEG & SLD & UCE \\ \midrule
        1 & 0.2090 & 0.0816 & 0.0144 \\ 
        100 & 0.8105 & 0.7530 & 0.0075 \\ \bottomrule
    \end{tabular}
    \label{tab:appd_multi}
\end{table}

\begin{table}[!t]
\caption{CLIP score (model: DPO-Diffusion)}
    \centering
    \resizebox{0.95\linewidth}{!}{\begin{tabular}{lllllll}
    \toprule
        & EMCID & ~ & ~ & SDD & ~ & ~ \\ \midrule
        Concept category & nudity & copyright & object & nudity & copyright & object \\
        Out-prompt retainability & 0.3320 & 0.3313 & 0.3314 & 0.3190 & 0.3129 & 0.2601 \\
        In-prompt retainability & 0.2929 & 0.2400 & 0.2838 & 0.2625 & 0.2388 & 0.2277 \\ \bottomrule
    \end{tabular}}
    \label{tab:appd_clip_score}
\end{table}

\begin{figure}[t]
    \centering
    \includegraphics[width=\linewidth]{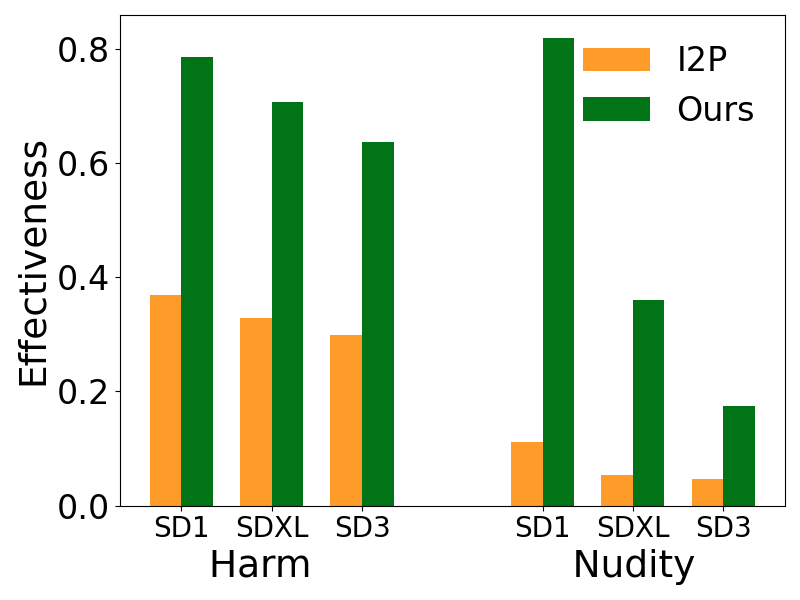}
    \caption{Effectiveness on SDXL and SD3.}
    \label{fig:effective}
\end{figure}

\begin{table}[t]
    \centering
    \caption{Detection Rate and Retainability on SD3}
    \resizebox{0.95\linewidth}{!}{\begin{tabular}{ccccccc}
    \toprule
        \textbf{Detect.} & Harm & Nudity & Celeb. & Copy. & Obj. & Art \\
    \midrule
        ORG & 0.637 & 0.174 & 0.980 & 1.000 & 0.980 & 0.275 \\
        NEG & 0.467 & 0.071 & 0.143 & 0.714 & 0.786 & 0.244 \\ 
        ESD & 0.628 & 0.138 & 0.393 & 1.000 & 0.643 & 0.246 \\ 
        SPM & 0.633 & 0.180 & 0.143 & 1.000 & 0.393 & 0.268 \\ 
        SDD & 0.635 & 0.131 & 0.679 & 0.321 & 0.214 & 0.273 \\ 
        SLD & 0.495 & 0.107 & 0.143 & 0.964 & 0.929 & 0.255 \\ 
    \bottomrule
    \toprule
        \textbf{Retain.} & Harm & Nudity & Celeb. & Copy. & Obj. & Art \\
    \midrule
        NEG & 0.274 & 0.286 & 0.273 & 0.271 & 0.298 & 0.314 \\
        ESD & 0.274 & 0.290 & 0.264 & 0.259 & 0.277 & 0.315 \\
        SPM & 0.260 & 0.290 & 0.248 & 0.252 & 0.269 & 0.315 \\
        SDD & 0.258 & 0.291 & 0.265 & 0.250 & 0.240 & 0.312 \\
        SLD & 0.273 & 0.290 & 0.266 & 0.264 & 0.296 & 0.315 \\
    \bottomrule
    \end{tabular}}
    \label{tab:sd3}
\end{table}

We also discuss the transferability of Six-CD and DVD as follows.

(1) Prompt effectiveness of Six-CD. In Fig.~\ref{fig:effective}, we show the effectiveness (i.e. $n/N$) of Harm and Nudity on SDXL and SD3. While our dataset generates fewer unwanted concepts on these models, it remains much more effective than other datasets like I2P. Specifically, its effectiveness is around three times as high as I2P for Nudity and twice for Harm.

(2) Results of removal and retainability. In Table~\ref{tab:sd3}, we provide results of 5 removal methods on SD3. We observe similar trends as in SD1. For example, prompts of specific concepts have higher effectiveness than general concepts. The detection of specific unwanted concepts are higher than 0.98 on ORG (i.e. models without concept removal). Another key observation, consistent with SD1, is that general concepts are more difficult to remove. Specifically, on general concepts, concept removal methods result in only a minimal reduction in detection rates compared to ORG.

\end{document}